\documentclass[letterpaper]{article} 
\usepackage{aaai25}  
\usepackage{times}  
\usepackage{helvet}  
\usepackage{courier}  
\usepackage[hyphens]{url}  
\usepackage{graphicx} 
\usepackage{natbib}  
\usepackage{caption} 
\usepackage{algorithm}
\usepackage{algorithmic}

\usepackage{newfloat}
\usepackage{listings}
\DeclareCaptionStyle{ruled}{labelfont=normalfont,labelsep=colon,strut=off} 
\floatstyle{ruled}
\newfloat{listing}{tb}{lst}{}
\floatname{listing}{Listing}
\usepackage{kotex}
\usepackage{mathtools}
\usepackage{xspace}
\usepackage{amsmath}
\usepackage{multirow}
\usepackage{amssymb}

\usepackage{subcaption}
\usepackage{makecell}
\usepackage{siunitx}
\usepackage{adjustbox}

\usepackage{hyperref} 

\newcommand{\set}[1]{\mathcal{#1}}

\providecommand{\sC}{\ensuremath{\set{C}}}
\providecommand{\sD}{\ensuremath{\set{D}}}
\providecommand{\sE}{\ensuremath{\set{E}}}

\providecommand{\sG}{\ensuremath{\set{G}}}

\providecommand{\sL}{\ensuremath{\set{L}}}
\providecommand{\sN}{\ensuremath{\set{N}}}

\providecommand{\sJ}{\ensuremath{\set{J}}}

\providecommand{\sT}{\ensuremath{\set{T}}}
\providecommand{\sV}{\ensuremath{\set{V}}}
\providecommand{\sX}{\ensuremath{\set{X}}}
\providecommand{\sY}{\ensuremath{\set{Y}}}

\providecommand{\sbP}{\ensuremath{\overline{\set{P}}}}
\providecommand{\stP}{\ensuremath{\widetilde{\set{P}}}}

\renewcommand{\vec}[1]{{\bf{#1}}}

\providecommand{\vb}{\ensuremath{\vec{b}}}

\providecommand{\vh}{\ensuremath{\vec{h}}}
\providecommand{\vr}{\ensuremath{\vec{r}}}
\providecommand{\vs}{\ensuremath{\vec{s}}}

\providecommand{\vx}{\ensuremath{\vec{x}}}

\providecommand{\vm}{\ensuremath{\vec{m}}}

\providecommand{\vq}{\ensuremath{\vec{q}}}

\providecommand{\vbb}{\ensuremath{\overline{\vec{b}}}}

\providecommand{\vbp}{\ensuremath{\overline{\vec{p}}}}
\providecommand{\vbr}{\ensuremath{\overline{\vec{r}}}}
\providecommand{\vbw}{\ensuremath{\overline{\vec{w}}}}

\providecommand{\vbz}{\ensuremath{\overline{\vec{z}}}}

\providecommand{\vtp}{\ensuremath{\widetilde{\vec{p}}}}
\providecommand{\vtr}{\ensuremath{\widetilde{\vec{r}}}}

\providecommand{\vtz}{\ensuremath{\widetilde{\vec{z}}}}

\providecommand{\vhp}{\ensuremath{\widehat{\vec{p}}}}
\providecommand{\vhr}{\ensuremath{\widehat{\vec{r}}}}
\providecommand{\vhx}{\ensuremath{\widehat{\vec{x}}}}
\providecommand{\vhz}{\ensuremath{\widehat{\vec{z}}}}

\newcommand{\mat}[1]{\mathbf{#1}}

\providecommand{\mD}{\ensuremath{\mat{D}}}

\providecommand{\mF}{\ensuremath{\mat{F}}}

\providecommand{\mR}{\ensuremath{\mat{R}}}

\providecommand{\mW}{\ensuremath{\mat{W}}}

\providecommand{\mZ}{\ensuremath{\mat{Z}}}

\providecommand{\mbR}{\ensuremath{\overline{\mat{R}}}}
\providecommand{\mbW}{\ensuremath{\overline{\mat{W}}}}
\providecommand{\mbZ}{\ensuremath{\overline{\mat{Z}}}}

\providecommand{\mtR}{\ensuremath{\widetilde{\mat{R}}}}

\providecommand{\mtZ}{\ensuremath{\widetilde{\mat{Z}}}}
\providecommand{\mhD}{\ensuremath{\widehat{\mat{D}}}}
\providecommand{\mhE}{\ensuremath{\widehat{\mat{E}}}}

\providecommand{\mhR}{\ensuremath{\widehat{\mat{R}}}}
\providecommand{\mhZ}{\ensuremath{\widehat{\mat{Z}}}}

\newcommand{\gcoporg}[0]{\textsl{GossipCop}\xspace}
\newcommand{\gcop}[0]{\textsl{GossipCop-S}\xspace}
\newcommand{\yelp}[0]{\textsl{YelpChi}\xspace}
\newcommand{\insr}[0]{\textsl{LifeIns}\xspace}

\newcommand{\ogbp}[0]{\textsl{OGB-Prod}\xspace}
\newcommand{\pubmed}[0]{\textsl{PubMed}\xspace}
\newcommand{\wikics}[0]{\textsl{Wiki-CS}\xspace}

\newcommand{\dgl}[0]{\textup{Deep Graph Library}\xspace}
\newcommand{\torch}[0]{\textup{PyTorch}\xspace}

\newcommand{\ours}[0]{\textup{MonTi}\xspace}

\newcommand{\clean}[0]{\texttt{\textup{Clean}}\xspace}
\newcommand{\gnia}[0]{\textup{G-NIA}\xspace}
\newcommand{\tdgia}[0]{\textup{TDGIA}\xspace}
\newcommand{\cluster}[0]{\textup{Cluster Attack}\xspace}
\newcommand{\gsquare}[0]{\textup{G$^{2}$A2C}\xspace}

\newcommand{\random}[0]{\textup{Random}\xspace}
\newcommand{\nettack}[0]{\textup{Nettack}\xspace}
\newcommand{\nipa}[0]{\textup{NIPA}\xspace}
\newcommand{\afgsm}[0]{\textup{AFGSM}\xspace}

\newcommand{\gcn}[0]{\textup{GCN}\xspace}
\newcommand{\gsage}[0]{\textup{GraphSAGE}\xspace}
\newcommand{\gat}[0]{\textup{GAT}\xspace}

\newcommand{\caregnn}[0]{\textup{CARE-GNN}\xspace}
\newcommand{\pcgnn}[0]{\textup{PC-GNN}\xspace}
\newcommand{\gaga}[0]{\textup{GAGA}\xspace}

\newcommand{\fone}[0]{F1-macro\xspace}
\newcommand{\auc}[0]{AUC\xspace}

\usepackage{pifont}
\newcommand{\cmark}{\ding{52}}

\newcolumntype{C}[1]{>{\centering\arraybackslash}p{#1}}

\title{Unveiling the Threat of Fraud Gangs to Graph Neural Networks:
\\ Multi-Target Graph Injection Attacks Against GNN-Based Fraud Detectors}
\author{
    Jinhyeok Choi, Heehyeon Kim, Joyce Jiyoung Whang\thanks{Corresponding author.}
}
\affiliations{
    School of Computing, KAIST\\
    \{cjh0507, heehyeon, jjwhang\}@kaist.ac.kr
}

\begin{document}

\maketitle

\begin{abstract}
Graph neural networks (GNNs) have emerged as an effective tool for fraud detection, identifying fraudulent users, and uncovering malicious behaviors. However, attacks against GNN-based fraud detectors and their risks have rarely been studied, thereby leaving potential threats unaddressed. Recent findings suggest that frauds are increasingly organized as gangs or groups. In this work, we design attack scenarios where fraud gangs aim to make their fraud nodes misclassified as benign by camouflaging their illicit activities in collusion. Based on these scenarios, we study adversarial attacks against GNN-based fraud detectors by simulating attacks of fraud gangs in three real-world fraud cases: spam reviews, fake news, and medical insurance frauds. We define these attacks as multi-target graph injection attacks and propose \ours, a transformer-based \underline{M}ulti-target \underline{on}e-\underline{T}ime graph \underline{i}njection attack model. \ours simultaneously generates attributes and edges of all attack nodes with a transformer encoder, capturing interdependencies between attributes and edges more effectively than most existing graph injection attack methods that generate these elements sequentially. Additionally, \ours adaptively allocates the degree budget for each attack node to explore diverse injection structures involving target, candidate, and attack nodes, unlike existing methods that fix the degree budget across all attack nodes. Experiments show that \ours outperforms the state-of-the-art graph injection attack methods on five real-world graphs.
\end{abstract}

%

\section{Introduction}
Recent endeavors have highlighted the power of Graph Neural Networks (GNNs) for fraud detection~\cite{upfd, get}. In fraud detection tasks, complex interactions of fraudsters can be effectively modeled using graphs, and frauds are typically represented as nodes corresponding to individuals with malicious intentions. GNN-based fraud detection methods aim to determine whether the nodes are fraudulent or benign. Meanwhile, it has been reported that adversarial attacks can cause GNNs to malfunction across various domains, including recommender systems and social network analysis~\cite{kgattack, anti, sui, toak, marl}. Those works consider vanilla GNNs as victim models, e.g., \gcn~\cite{gcn}, \gsage~\cite{gsage}, and \gat~\cite{gat}. On the other hand, various tailored GNNs for fraud detection have recently been developed to filter the camouflaged fraudsters, such as CARE-GNN~\cite{caregnn}, PC-GNN~\cite{pcgnn}, and GAGA~\cite{gaga}. However, vulnerabilities of these GNN-based fraud detectors to adversarial attacks remain unexplored.

\begin{figure}
\centering
\includegraphics[width=0.91\columnwidth]{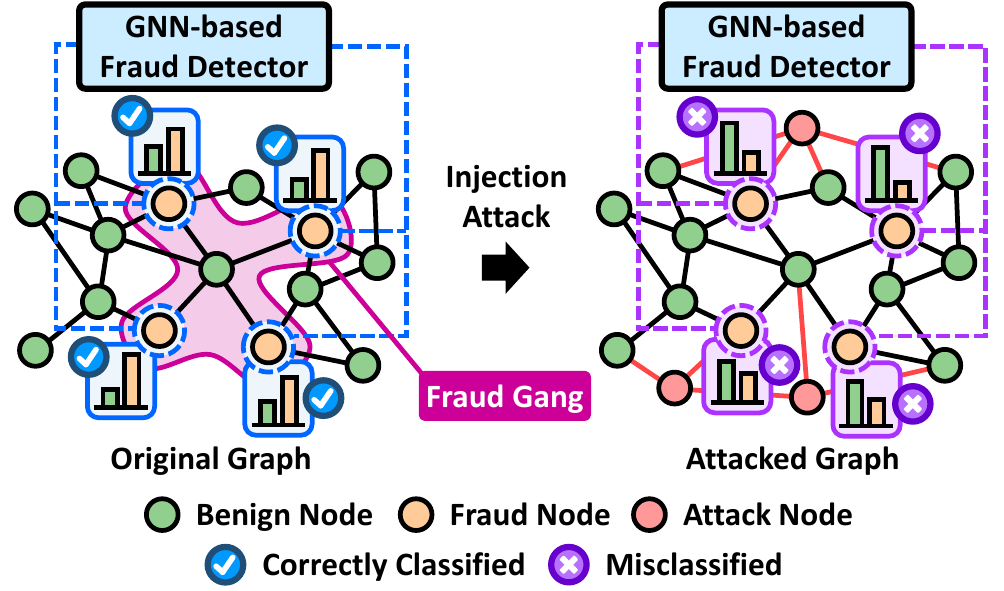}
\caption{An example of a multi-target graph injection attack against a GNN-based fraud detector: a fraud gang injects attack nodes to induce misclassification of their fraud nodes.}
\label{fig:example}
\end{figure}

It has been recently observed that frauds are increasingly organized into gangs or groups exhibiting collusive patterns to carry out fraudulent activities more effectively with reduced risk~\cite{acd, gfdn, rdpgl}. For example, in the medical insurance domain, fraudsters may collaborate with doctors or insurance agents to obtain fake diagnoses. On online review platforms, fraudsters could create multiple fake reviews using different IDs. On social media platforms, fraudsters can spread misinformation by using multiple fake accounts. We design these three attack scenarios where fraud gangs attack GNN-based fraud detectors to make them misclassify the fraud nodes as benign. Furthermore, to simulate the scenarios, we create datasets and target sets that consist of fraud nodes grouped based on metadata or relations in real-world graphs.

In this work, the adversarial attack on GNN-based fraud detectors by fraud gangs is defined as a multi-target graph injection attack. We adopt a graph injection attack, as it is more feasible than a graph modification attack, which requires privileged access to alter existing structures~\cite{hao}. As illustrated in Figure~\ref{fig:example}, in our attack scenarios, a fraud gang attempts to deceive the detector by injecting malicious nodes (i.e., attack nodes) into the original graph, causing their fraud nodes to be misclassified by increasing the scores of being benign. We consider a black-box evasion attack, where the attacker can access only the original graph, partial labels, and a surrogate model, and the attack occurs during the victim model's inference phase~\cite{survey2}.

Existing methods~\cite{gnia, tdgia, cluster} have focused on adversarial attacks for randomly grouped target nodes. However, fraud nodes are often organized into gangs to camouflage their illicit activities. While the existing methods have investigated attacks where vanilla GNNs are employed as victim models, GNN-based fraud detectors are tailored to address heterophily induced by fraud nodes, making them more difficult to attack than vanilla GNNs. To overcome the limitations of these existing methods in our scenario, we propose \ours, a transformer-based \underline{M}ulti-target \underline{on}e-\underline{T}ime graph \underline{i}njection attack model.

\ours significantly differs from the existing graph injection attack methods in several aspects, as summarized in Table~\ref{tb:difference}. The existing methods, such as \tdgia~\cite{tdgia}, \cluster~\cite{cluster}, and \gsquare~\cite{g2a2c}, inject multiple attack nodes sequentially, fixing the graph structure at each step. Those approaches fix the degree budget across all attack nodes due to the lack of information about future steps, limiting their flexibility and efficiency. In contrast, \ours injects all attack nodes at once, adaptively assigning the degree budget for each attack node to explore diverse structures among target, candidate, and attack nodes. Furthermore, the existing methods, including \gnia~\cite{gnia}, overlook interactions within target nodes and among attack nodes, which are crucial for modeling the intricate relationships within a fraud gang and attack nodes. Additionally, the existing methods sequentially generate attributes and edges of attack nodes, only considering a one-way dependency, which fails to model collusive patterns of fraud gangs. On the other hand, \ours employs a transformer encoder to effectively capture interdependencies within target and attack nodes, and between attributes and edges. Our contributions are summarized as follows:
\begin{itemize}
    \item To the best of our knowledge, our work is the first study to investigate the vulnerabilities of GNN-based fraud detectors and also the first study on graph injection attacks for multiple target nodes organized by groups.
    \item We propose a graph injection attack method \ours, which generates attributes and edges of all attack nodes at once via the adversarial structure encoding transformer.
    \item \ours can explore adversarial injection structures comprehensively by generating adversarial edges with adaptive degree budget allocation for each attack node.
    \item Experimental results demonstrate that \ours substantially outperforms the state-of-the-art graph injection attack methods on five real-world graphs.\footnote{Our datasets and codes are available at \href{https://github.com/bdi-lab/MonTi}{https://github.com/bdi-lab/MonTi}; the full paper is available at \href{https://bdi-lab.kaist.ac.kr}{https://bdi-lab.kaist.ac.kr}.}
\end{itemize}

\begin{table}[t]
\small
\renewcommand{\arraystretch}{1.05}
\setlength{\tabcolsep}{0.05em}
\centering

\begin{tabular}{c|C{1.95em}C{1.95em}|C{1.95em}C{1.95em}|C{1.95em}C{1.95em}|C{1.95em}C{1.95em}|C{1.95em}C{1.95em}}
\Xhline{2\arrayrulewidth}
 & \multicolumn{2}{C{4.06em}|}{Victim} & \multicolumn{2}{C{4.06em}|}{Multi-Tar.} & \multicolumn{2}{C{4.06em}|}{Multi-Inj.} & \multicolumn{2}{C{4.06em}|}{Budget} & \multicolumn{2}{C{4.06em}}{Attr-Edge} \\ \cline{2-9}
 & Van & Frd & Rnd & Frd & Seq & One & Fix & Ada & \multicolumn{2}{C{4.06em}}{Interdep.} \\
\Xhline{\arrayrulewidth}
\gnia        & \cmark & & \cmark & & & & \cmark & & & \\
\tdgia       & \cmark & & \cmark & & \cmark & & \cmark & & &  \\
Cluster Atk. & \cmark & & \cmark & & \cmark & & \cmark & & &  \\
\gsquare     & \cmark & & & & \cmark & & \cmark & & & \\
\Xhline{\arrayrulewidth}
\ours        & \cmark & \cmark & \cmark & \cmark & & \cmark & & \cmark & \multicolumn{2}{c}{\cmark} \\
\Xhline{2\arrayrulewidth}
\end{tabular}

\caption{Comparison of graph injection attack methods. We mark whether victim models are vanilla GNNs (Van) or GNN-based fraud detectors (Frd); target nodes are randomly grouped (Rnd) or fraud gangs (Frd); multiple attack nodes are injected sequentially (Seq) or at once (One); degree budget per attack node is fixed (Fix) or adaptively assigned (Ada); attribute-edge interdependency is considered or not.}

\label{tb:difference}
\end{table}

\section{Related Work}
\paragraph{\normalfont{\textbf{Graph-based Fraud Detection}}}
The majority of graph-based fraud detection methods adopt GNNs and define this task as a problem of classifying nodes into two categories: fraud or benign~\cite{aognn, h2fd, amnet, dignn, dagnn}. In fraud graphs, fraud nodes are much fewer than benign nodes, resulting in a class imbalance problem. In addition, fraud nodes tend to connect with or imitate the attributes of benign nodes to hide their malicious activities. Such behavior introduces heterophily~\cite{hetero}, which causes vanilla GNNs to fail in detecting fraud nodes. To address the class imbalance and heterophily in fraud graphs, recent approaches have introduced learnable neighborhood samplers~\cite{caregnn, pcgnn}, utilized Beta wavelet transform~\cite{bwgnn, split} and proposed group aggregation strategies~\cite{gaga, dgagnn}. Despite the recent advancements, vulnerabilities of GNN-based fraud detectors to adversarial attacks have not yet been studied.

\paragraph{\normalfont{\textbf{Adversarial Attacks on Graphs}}}
It has been shown that even a few adversarial modifications on graphs can significantly degrade the performance of GNNs~\cite{nettack, gp, metattack, rewatt, infmax, spac}. Graph injection attacks have emerged focusing on more practical settings, where attackers inject new malicious nodes instead of modifying existing nodes and edges~\cite{afgsm, nipa}. Recent developments include learning attribute and edge generators to target unseen nodes~\cite{gnia}, introducing defective edge selection strategies~\cite{tdgia}, regarding graph injection attacks as clustering tasks~\cite{cluster}, and employing the advantage actor-critic framework~\cite{g2a2c}. More detailed discussions on related work are provided in Appendix~\ref{sec:app_rel_work}.

\begin{figure*}[t]
\centering
\includegraphics[width=2.11\columnwidth]{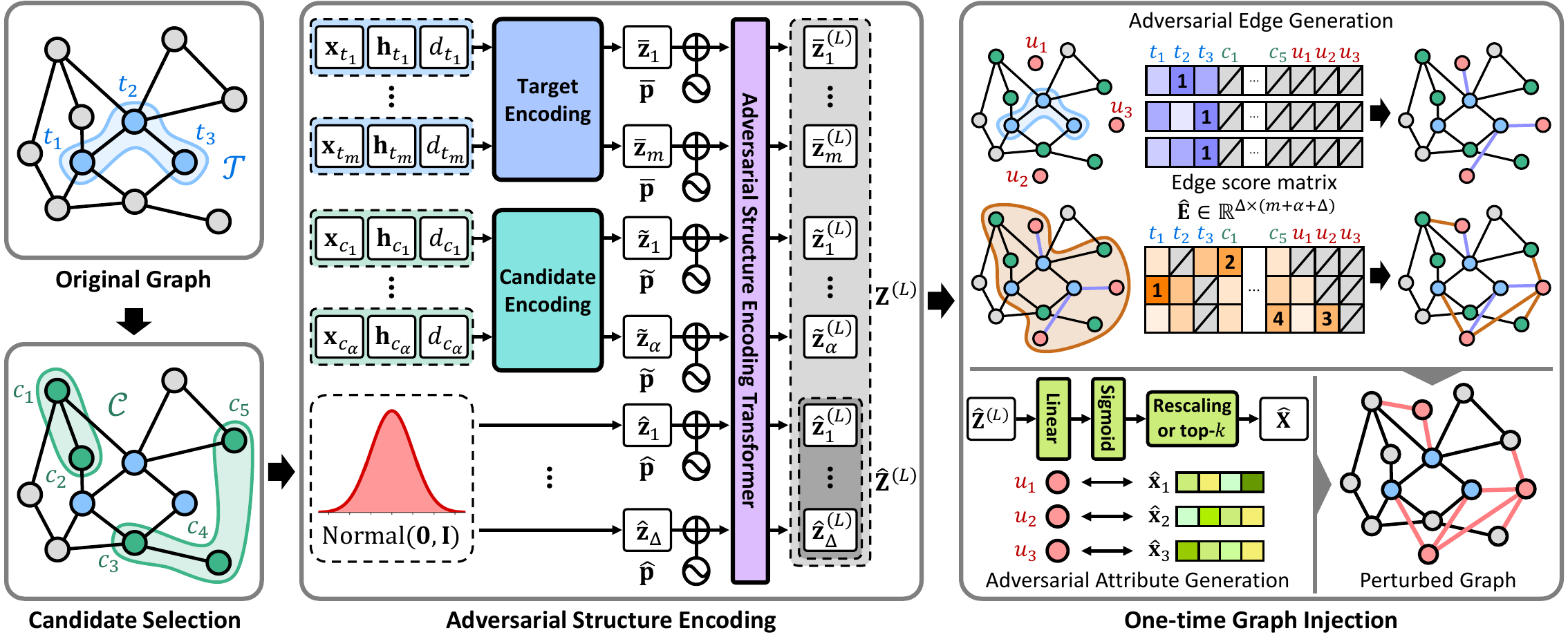}
\caption{Overview of \ours. Given the original graph and the target set, \ours selects candidate nodes among $K$-hop neighbors of target nodes utilizing a learnable scoring function. Then, \ours calculates the intermediate node representations of target, candidate, and attack nodes by the adversarial structure encoding transformer. Based on those intermediate representations, adversarial attributes and edges for injection are simultaneously generated.}
\label{fig:ours}
\end{figure*}

\section{Problem Definition}
An undirected attributed graph is defined as $G=(\sV,\sE,\sX)$ where $\sV$ is a set of $n$ nodes, $\sE\subset\sV\times\sV$ is a set of edges, and $\sX$ is a set of $D$-dimensional node attribute vectors. For a node $v\in\sV$, we represent its attribute vector as $\vx_v\in\sX$ and its label as $y_v \in \{0, 1\}$ where $y_v=1$ indicates $v$ is a fraud and $y_v=0$ indicates $v$ is benign. $\sY = \{y_v \,|\, v \in \sV \}$ denotes a set of node labels. We represent a GNN-based fraud detector as $f_\theta(\cdot)$ where $\theta$ indicates learnable parameters. The fraud score vector $\vs_v\in\mathbb{R}^{2}$ and the predicted label $\hat{y}_v \in \{0, 1\}$ are calculated by $\vs_v = f_\theta(G, v) = \text{MLP}\left( \Phi(G, v) \right)$ and $\hat{y}_v = {\arg\max}_{i}\;{s_{v,i}}$ where MLP is a multi-layer perceptron, $\Phi(\cdot)$ denotes a GNN encoder and $s_{v,i}$ denotes the $i$-th element of $\vs_{v}$. We formulate the objective function of the fraud detector as $\max_{\theta}{\sum_{v\in\sV} \mathbb{I}\,(\hat{y}_v = y_v)}$ where $\mathbb{I}\,(\cdot)$ is an indicator function. Most GNN-based fraud detectors utilize label information based on domain-specific knowledge, employing (semi-)supervised approaches~\cite{caregnn, bond}. As the first study of adversarial attacks against GNN-based fraud detectors, we assume that graphs are single-relational, in line with existing studies of adversarial attacks on graphs~\cite{gnia, tdgia, cluster, g2a2c}. 

A graph injection attack injects attack nodes $\sV_{\text{in}}$ with attributes $\sX_{\text{in}}$ and adversarial edges $\sE_{\text{in}} \subset (\sV' \times \sV') \setminus (\sV \times \sV)$ into the original graph $G=(\sV,\sE,\sX)$ where $\sV'=\sV\cup\sV_{\text{in}}$. We define the perturbed graph as $G'=(\sV', \sE', \sX')$ where $\sE'=\sE\cup\sE_{\text{in}}$ and $\sX'=\sX\cup\sX_{\text{in}}$. The multi-target graph injection attack against a GNN-based fraud detector aims to make fraud nodes in the target set $\sT\subset\sV$ misclassified. For a node $v \in \sV$, let $\vs'_v = f_{\theta^{*}}(G', v)$ where $\theta^{*} = {\arg\min}_{\theta}\;{\sL_\text{train}(f_\theta, G, \sD)}$, $\sL_\text{train}$ is a training loss of $f_\theta(\cdot)$, and $\sD \subset \sY$ is a node label set for training. The objective function of the multi-target graph injection attack is:
\begin{equation}
    \adjustbox{scale=1.00}{$\displaystyle\min_{G'} \sum_{t \in \mathcal{T}} \mathbb{I}\left(\arg\max_i s'_{t,i} = y_t\right) \, \text{st.} \, |\mathcal{V}_{\text{in}}| \leq \Delta, |\mathcal{E}_{\text{in}}| \leq \eta$}
\end{equation}
where $\Delta$ and $\eta$ denote node and edge budgets, respectively.

\section{Multi-target One-time Graph Injection Attack}
We propose \ours (Figure~\ref{fig:ours}), a graph injection attack model with a transformer-based architecture. \ours utilizes self-attention to capture interactions among target, candidate, and attack nodes, simultaneously generating attributes and edges of attack nodes to reflect their interdependencies. In addition, by generating edges of all attack nodes at once, \ours enables flexible budget allocation for each node.

\subsection{Adversarial Structure Encoding}
Three types of contexts can affect multi-target graph injection attacks: target nodes $\sT=\{t_1, \cdots, t_m\}$, attack nodes $\sV_{\text{in}}=\{u_1, \cdots, u_\Delta\}$, and candidate nodes. We define $\sN^{(K)} \subset \sV$ as a set of $K$-hop neighbors of the target nodes, excluding the target nodes themselves. From $\sN^{(K)}$, we select $\alpha$ nodes to form a set of candidate nodes $\sC=\{c_1, \cdots, c_\alpha\} \subset \sN^{(K)}$. The neighborhood of target nodes provides information on how to propagate malicious messages within a GNN-based architecture. \ours employs multi-head self-attention to learn intermediate representations for nodes in $\sT$, $\sV_{\text{in}}$, and $\sC$ to capture interactions among these contexts, distinguishing their respective roles.

\subsubsection{Candidate Selection}
The size of $\sN^{(K)}$ can drastically increase depending on the target nodes, making it computationally inefficient and challenging to find the optimal $G'$. Therefore, when the number of possible candidate nodes exceeds the threshold $n_c$, \ours selects candidate nodes to narrow the search space with a learnable scoring function $\sJ$; otherwise, all nodes are considered as candidate nodes. To incorporate the class distribution learned by GNN architectures, \ours utilizes a surrogate GNN model pretrained on the original graph. \ours calculates candidate scores for all $v \in \sN^{(K)}$ by integrating node attributes and graph topology with those of the target nodes using the surrogate GNN:
\begin{equation}
    \adjustbox{scale=0.95}{$\displaystyle\sJ(G,v) = \text{MLP}\left( \sigma \left( [\vq_v \,||\, \vm_v \,||\, \vh_v \,||\, \vh_{\sT}] \right) \right)$}
\end{equation}
where $\vq_v = \text{MLP}(\vx_v) \in \mathbb{R}^{D_H}$, $D_H$ is the dimension of node representation, $\vm_v = \text{MLP}([d_v||\beta_v]) \in \mathbb{R}^{D_H}$, $d_v$ is the degree of node $v$, $\beta_v$ is the number of target nodes directly connected to $v$, $\vh_{v}=\Phi^{*}(G,v)\in\mathbb{R}^{D_H}$, $\Phi^{*}$ is a pretrained GNN encoder of the surrogate model, $\vh_{\sT} = \text{READOUT}(\{\vh_{t} \,|\, t \in \sT \})\in \mathbb{R}^{D_H}$, $\text{READOUT}(\cdot)$ is a pooling function, $||$ is a horizontal concatenation, and $\sigma$ is the activation function. We adopt the mean operation as a pooling function. Based on the calculated scores, the top-$n_c$ nodes are selected as candidate nodes. To solve the optimization problems of discrete selection in \ours, we adopt the Gumbel-Top-$k$ technique~\cite{gsoft, gtopk} coupled with the straight-through estimator~\cite{straight}.

\subsubsection{Target and Candidate Encoding}
To exploit the structural information of target and candidate nodes, we introduce learnable encodings with three different factors of each node $v$ related to the attack: raw attributes $\vx_{v}$, degree $d_{v}$, and the GNN representation $\vh_{v}$. We employ the raw attributes to explicitly take into account the node itself. The node degree is utilized to indirectly measure the susceptibility of the node to the change of its neighbors. We use the GNN representation to capture an inherent class distribution learned by the surrogate GNN and topological information of the node. With a learnable target encoding function $\sbP$, a target encoding $\vbz_{i} = \sbP(\vx_{t_i}, d_{t_i}, \vh_{t_i})$ for a target node $t_i \in \mathcal{T}$ is computed by projecting and integrating all three factors:
\begin{equation}
    \adjustbox{scale=0.95}{$\displaystyle\vbz_{i} = \text{MLP}\left( \sigma([(\mbW\vx_{t_i}+\vbb_{1}) \,||\, (\vbw d_{t_i}+\vbb_{2}) \,||\, \vh_{t_i}] ) \right)$}
\end{equation}
where \text{MLP} transforms the input into $\mathbb{R}^{D_H}$, $\mbW\in\mathbb{R}^{{D_H}\times{D}}$, $\vbw\in\mathbb{R}^{D_H}$, $\vbb_{1}\in\mathbb{R}^{D_H}$, $\vbb_{2}\in\mathbb{R}^{D_H}$. Similarly, we compute a candidate encoding $\vtz_{i}=\stP(\vx_{c_i}, d_{c_i}, \vh_{c_i})$ for a candidate node $c_i \in \mathcal{C}$ where $\stP$ denotes a learnable candidate encoding function. Note that the parameters of $\sbP$ and $\stP$ are distinct.

\subsubsection{Adversarial Structure Encoding Transformer}
We propose to adopt the transformer~\cite{att} encoder to learn the intermediate node representations of the target, candidate, and attack nodes for adversarial attribute and edge generation. Each attack node $u_i \in \sV_{\text{in}}$ is initialized with the representation $\vhz_i \in \mathbb{R}^{D_H}$ sampled from a standard Gaussian distribution. The input sequence for the adversarial structure encoding transformer is defined as ${\mZ}^{(0)} = [ \mbZ\,^{(0)} || \mtZ^{(0)} || \mhZ^{(0)}] = [\, \vbz_1 \cdots \vbz_m || \vtz_1 \cdots \vtz_{\alpha} || \vhz_1  \cdots \vhz_{\Delta} ]$ where $\mbZ\,^{(0)}$, $\mtZ^{(0)}$, and $\mhZ^{(0)}$ denote the input sequence corresponding to the target, candidate, and attack nodes, respectively. We then add learnable positional encodings $\vbp, \; \vtp, \; \vhp \in \mathbb{R}^{D_H}$ to the corresponding node representations. Subsequently, ${\mZ}^{(0)}$ is processed through an $L$-layer transformer encoder with $n_h$ heads to obtain the final representations ${\mZ^{(L)}} = [\mbZ\,^{(L)} || \mtZ^{(L)} || \mhZ^{(L)}]$. To mitigate noise from attack nodes whose edges are not formed yet, we optionally mask the attack nodes when calculating target and candidate node representations, treating this masking as a hyperparameter.

\subsection{Adversarial Attribute Generation}
\label{sec:attrgen}
To generate the attributes of attack nodes, we project ${\mhZ}^{(L)}$ into $D$-dimensional space: $\mF = \text{sigmoid}(\mW_{a}{\mhZ}^{(L)} + \vb_{a}) \in \mathbb{R}^{{D_H}\times{\Delta}}$ where $\mW_{a} \in \mathbb{R}^{{D}\times{D_H}}$ and $\vb_{a} \in \mathbb{R}^{D}$. Depending on whether the raw attributes are continuous or discrete, $\mF$ is transformed into the malicious attributes in different ways. 

For continuous attributes, we rescale $\mF$ by using the min-max vectors of raw attributes $\vx_{\text{min}}, \vx_{\text{max}}\in\mathbb{R}^{D}$ derived from the original graph. The adversarial attribute vector $\vhx_i$ for the attack node $u_i \in \sV_{\text{in}}$ is computed by $\vhx_i = \mF_i\odot (\vx_{\text{max}}-\vx_{\text{min}}) + \vx_{\text{min}}$ where $\mF_i \in \mathbb{R}^{D}$ denotes the $i$-th column vector of $\mF$, and $\odot$ denotes the element-wise product operator.

We adopt the Gumbel-Top-$k$ technique to optimize discrete choices, such as candidate selection, discrete attribute generation, and edge generation. Following \cite{gnia, grsp}, we additionally utilize the exploration parameter $\epsilon$ to control the randomness. The Gumbel-Softmax for the discrete attribute generation is defined as:
\begin{equation}
\label{eqn:gsoft}
    \adjustbox{scale=0.92}{$\displaystyle\text{Gumbel-Softmax}(\mF_i, \epsilon)_j = \frac{\text{exp}\left((F_{ij}+\epsilon N_j) / \tau \right)} {\sum_{j'=1}^{D} \text{exp} \left( (F_{ij'}+\epsilon N_{j'} / \tau) \right)}$}
\end{equation}
where $F_{ij}$ is the $j$-th element of $\mF_i$, $N_j = -\text{log}(-\text{log}(U_j))$, $U_j\sim\text{Uniform}(0,1)$ is a Uniform random variable, and $\tau$ is the temperature parameter. The Gumbel-Top-$k$ function $\sG^a$ for discrete attribute generation is defined as:
\begin{equation}
    \adjustbox{scale=0.95}{$\displaystyle
    \sG^a(\mF_i) = \underset{j}{\arg\text{top-}k}\;{ \text{Gumbel-Softmax}(\mF_i, \epsilon)_j}$}
\end{equation}
where $\arg\text{top-}k$ function returns the indices corresponding to the $k$ highest values of the input. Here, we set $k$ as $\lambda$, the average count of non-zero values in the node attributes of the entire graph. Finally, the $j$-th element of $\vhx_i$ becomes $\widehat{x}_{ij} = \mathbb{I} \left( j \in \sG^a(\mF_i) \right)$. We employ the straight-through estimator to optimize the discrete selection processes, allowing for gradient-based optimization with discrete elements.

\subsection{Adversarial Edge Generation}
\label{sec:edgegen}
To generate all adversarial edges at once within the edge budget $\eta$, we construct an edge score matrix and apply the Gumbel-Top-$k$. First, we project ${\mZ}^{(L)}$ into edge score space:
\begin{equation}
    \adjustbox{scale=0.95}{$\displaystyle
    \begin{aligned}
        \mR &= \mW_{e}{\mZ}^{(L)} + \vb_{e} = [\, \mbR \,||\, \mtR \,||\, \mhR\,] \\
        &= [\, \vbr_1 \; \cdots \; \vbr_m \;||\; \vtr_1 \; \cdots \; \vtr_{\alpha} \;||\; \vhr_1 \; \cdots \; \vhr_{\Delta} ] \\
        &= [\, \vr_1 \; \cdots \; \vr_M ]
    \end{aligned}$}
\end{equation}
where $\mW_{e}\in\mathbb{R}^{{D_H}\times{D_H}}$, $\vb_{e}\in\mathbb{R}^{D_H}$, $\mbR \in \mathbb{R}^{D_H \times m}$, $\mtR \in \mathbb{R}^{D_H \times \alpha}$, $\mhR \in \mathbb{R}^{D_H \times \Delta}$, $m$ is the number of target nodes, $\alpha$ is the number of candidate nodes, $\Delta$ is a node budget, and $M=m+\alpha+\Delta$. We define the edge score $e_{ij}$ between an attack node $u_i \in \sV_{\text{in}}$ and a node $v_j \in \sT \cup \sC \cup \sV_{\text{in}}$ as the cosine similarity between their representations. Then, the edge score matrix $\mhE \in  \mathbb{R}^{\Delta \times M}$ can be written as $\mhE = \mhD^{-1} \mhR^{\top} \mR \mD^{-1}$ where $\mD=\text{diag}(||\vr_1||, \cdots, ||\vr_M|| )$, and $\mhD=\text{diag}( ||\,\vhr_1||, \cdots, ||\,\vhr_\Delta|| )$. To guarantee that every attack node is directly connected to the original graph, we generate at least a single edge for each attack node to one of the target nodes by applying Gumbel-Top-$k$ with $k=1$. For all remaining possible edges, we apply Gumbel-Top-$k$ across the entire $\mhE$ with $k=\eta-\Delta$. Note that edge scores corresponding to self-loops and duplicated edges are masked.

\subsection{Training of \ours}
Following the previous works in graph injection attacks~\cite{gnia, cluster}, we define the loss function for \ours based on C\&W loss~\cite{loss}:
\begin{equation}
    \adjustbox{scale=1.0}{$\displaystyle\min_{G'} \; \sL (f_{\theta^{*}}, G', \sT) = \frac{1}{\lvert\sT\rvert}  {\sum_{t\in\sT}} \max\big( s'_{t, 1} - s'_{t, 0}\, ,\;0 \big)$}
\end{equation}
where $\vs'_t=f_{\theta^{*}}(G', t) \in \mathbb{R}^2$, and $s'_{t, i}$ denotes the $i$-th element of $\vs'_t$. We focus on increasing normal scores and decreasing fraud scores of target nodes to align with our attack scenarios. In line with our emphasis on black-box attack settings, the loss is calculated using a surrogate model.

\section{Experiments}
\label{sec:exp}
We compare \ours with the state-of-the-art graph injection attack baselines on five real-world graphs.

\paragraph{\normalfont{\textbf{Datasets}}}
\label{paragraph:Datasets}
Our experiments on multi-target graph injection attacks cover three real-world datasets: \gcop, \yelp, and \insr. \gcop and \yelp have continuous node attributes, whereas \insr contains discrete ones. Using \gcoporg~\cite{fnewsnet}, which includes news articles and their Twitter engagements~\cite{fnewsnet}, we create \gcop by following~\cite{gossip}, linking articles tweeted by the same multiple users. \yelp~\cite{yelp, fake} is a review graph where nodes represent reviews. \insr is a medical insurance graph based on real-world data provided by an anonymous insurance company. In \insr, nodes correspond to claims, and edges represent relationships predefined by domain experts. In experiments, we use $p\%$ of nodes as training sets, setting $p=40$ for \gcop and \yelp, and $p=10$ for \insr. The remaining nodes are split into validation and test sets with a ratio of 1:2, following the conventional setting in the GNN-based fraud detection~\cite{bwgnn}. We create the training, validation, and test target sets with fraud nodes belonging to each split. Each target set represents a fraud gang organized based on metadata or relations in each dataset. The statistic of datasets for multi-target attacks is summarized in Table~\ref{tb:data}. More detailed descriptions of the datasets are in Appendix~\ref{sec:app_data}.

Although \ours is designed for multi-target attacks, we present benchmark results for single-target attacks on \ogbp~\cite{ogbp} and \pubmed~\cite{pubmed} in Appendix E.1. Despite not being specifically designed for single-target attacks, \ours achieves the highest misclassification rates on both datasets. This demonstrates the effectiveness of \ours's transformer-based architecture, which captures intricate interactions around a target node via adversarial structure encoding.

\paragraph{\normalfont{\textbf{Budgets}}}
Due to the diverse sizes and substructures of target sets, node and edge budgets should be allocated according to the characteristics of each target set. We also impose limits on the budgets since excessively large budgets can lead to highly noticeable and easy attacks. The node budget $\Delta$ for each target set is defined as $\Delta=\max( \big\lfloor \rho \,\cdot\, \min(B, \, \overline{B} \,) + 0.5 \big\rfloor, \, 1)$ where $\rho$ is a parameter to control node budgets, $B = \lvert \sN^{(1)} \cup \sT \rvert$, and $\overline{B}$ is the average value of $B$ across all target sets within the dataset. The edge budget $\eta$ for each target set is calculated as $\eta=\Delta \,\cdot\, \max(\big\lfloor \min(d_\sT,\; \xi\cdot\overline{d} \,) + 0.5 \,\big\rfloor, \, 1)$ where $d_\sT$ is the average node degree of the target set, $\xi$ is a parameter to control edge budgets, and $\overline{d}$ is the average node degree of all nodes in the graph. Unless specifically stated otherwise, we set $\rho=0.05, \, \xi=0.1$ for \gcop, $\rho=0.05, \, \xi=0.5$ for \yelp, and $\rho=0.2, \, \xi=0.5$ for \insr. A detailed explanation of the rationale behind defining the budgets and analysis of the effect of $\rho$ and $\xi$ is provided in Appendix~\ref{subsec:app_budget}.

\begin{table}[t]
\small
\renewcommand{\arraystretch}{1.05}
\setlength{\tabcolsep}{0.25em}
\centering

\begin{tabular}{c|cccccc}
\Xhline{2\arrayrulewidth}
& $|\sV|$ & \#Frauds & \#Target Sets & $|\sE|$ & $D$ \\
\Xhline{\arrayrulewidth}
\gcop & 16,488 & 3,898 & 2,438 & 3,865,058 & 768 \\
\yelp & 45,900 & 6,656 & 1,435 & 3,846,910 & 32 \\
\insr & 122,792 & 1,264 & 380 & 912,833 & 1,611 \\
\Xhline{2\arrayrulewidth}
\end{tabular}

\caption{Statistic of datasets for multi-target attacks.}
\label{tb:data}

\end{table}

\begin{table*}[t]
\small
\renewcommand{\arraystretch}{1.05}
\setlength{\tabcolsep}{0.6em}

\centering

\begin{tabular}{cc|cccccc}
    \Xhline{2\arrayrulewidth}
    & & \multicolumn{6}{c}{Surrogate / Victim Model} \\ 
    Dataset & Attack Method & \gcn & \gsage & \gat & \caregnn & \pcgnn & \gaga \\ 
    \Xhline{\arrayrulewidth}
    \multicolumn{1}{c}{\multirow{6}{*}{\gcop}} & \clean & 46.70 & 26.04 & 11.29 & 48.02 & 55.62 & 21.68 \\
    \multicolumn{1}{c}{} & \gnia & \underline{75.12$\pm$0.11} & \underline{67.70$\pm$2.05} & \underline{63.21$\pm$3.33} & \underline{59.96$\pm$2.89} & \underline{62.60$\pm$1.53} & \underline{25.69$\pm$1.44} \\
    \multicolumn{1}{c}{} & \tdgia & 42.93$\pm$0.26 & 41.07$\pm$0.47 & 24.70$\pm$0.62 & 57.49$\pm$0.11 & 62.24$\pm$0.12 & 22.09$\pm$0.05 \\
    \multicolumn{1}{c}{} & \cluster & 43.67$\pm$0.21 & 39.89$\pm$0.49 & 24.88$\pm$0.33 & 57.12$\pm$0.17 & 61.83$\pm$0.15 & 22.09$\pm$0.05 \\
    \multicolumn{1}{c}{} & \gsquare & OOM & OOM & OOM & OOM & OOM & OOM \\
    \multicolumn{1}{c}{} & \ours & \textbf{92.60$\pm$0.34} & \textbf{97.05$\pm$0.15} & \textbf{94.30$\pm$2.55} & \textbf{90.15$\pm$0.44} & \textbf{90.12$\pm$0.17} & \textbf{46.94$\pm$1.48} \\
    \Xhline{\arrayrulewidth}
    \multicolumn{1}{c}{\multirow{6}{*}{\yelp}} & \clean & 87.14 & 43.81 & 35.12 & 29.79 & 59.13 & 28.00 \\
    \multicolumn{1}{c}{} & \gnia & \underline{90.93$\pm$0.90} & \underline{64.56$\pm$1.34} & \underline{55.51$\pm$11.72} & \textbf{32.45$\pm$1.13} & \underline{63.18$\pm$1.71} & \underline{31.08$\pm$0.43} \\
    \multicolumn{1}{c}{} & \tdgia & 86.87$\pm$0.10 & 43.28$\pm$0.23 & 46.95$\pm$1.14 & 30.66$\pm$0.24 & 58.01$\pm$0.21 & 28.03$\pm$0.02 \\
    \multicolumn{1}{c}{} & \cluster & 86.97$\pm$0.04 & 43.72$\pm$0.05 & 45.16$\pm$0.80 & 30.13$\pm$0.05 & 58.34$\pm$0.15 & 28.02$\pm$0.05 \\
    \multicolumn{1}{c}{} & \gsquare & OOM & OOM & OOM & OOM & OOM & OOM \\
    \multicolumn{1}{c}{} & \ours & \textbf{92.23$\pm$0.95} & \textbf{65.31$\pm$1.19} & \textbf{93.27$\pm$7.84} & \underline{31.92$\pm$0.53} & \textbf{69.93$\pm$1.31} & \textbf{37.66$\pm$0.92} \\
    \Xhline{\arrayrulewidth}
    \multicolumn{1}{c}{\multirow{6}{*}{\insr}} & \clean & 27.72 & 13.70 & 16.75 & 16.42 & 16.17 & 15.68 \\
    \multicolumn{1}{c}{} & \gnia & 33.50$\pm$4.84 & 13.20$\pm$1.02 & 16.50$\pm$0.79 & 16.09$\pm$0.43 & 16.09$\pm$0.47 & \underline{17.44$\pm$0.23} \\
    \multicolumn{1}{c}{} & \tdgia & \underline{83.28$\pm$0.17} & \underline{37.80$\pm$0.12} & \underline{96.60$\pm$0.59} & \underline{18.05$\pm$0.19} & 17.90$\pm$0.10 & 16.87$\pm$0.13 \\
    \multicolumn{1}{c}{} & \cluster & N/A & N/A & N/A & N/A & N/A & N/A \\
    \multicolumn{1}{c}{} & \gsquare & 45.05$\pm$1.82 & 13.00$\pm$0.05 & 35.85$\pm$2.02 & 17.24$\pm$0.00 & \underline{20.08$\pm$0.04} & OOM \\
    \multicolumn{1}{c}{} & \ours & \textbf{99.47$\pm$0.31} & \textbf{60.97$\pm$0.97} & \textbf{100.00$\pm$0.00} & \textbf{26.80$\pm$4.29} & \textbf{20.64$\pm$0.30} & \textbf{35.03$\pm$1.54} \\
    \Xhline{2\arrayrulewidth}
\end{tabular}

\caption{Multi-target attack performance on \gcop, \yelp, and \insr where the types of surrogate and victim models are the same. We report misclassification rates (\%). OOM: Out of Memory. N/A: Not completed in 5 days.}
\label{tb:mainexp1}

\end{table*}

\begin{table*}[t]
\small
\renewcommand{\arraystretch}{1.05}

\setlength{\tabcolsep}{0.25em}

\centering
\begin{tabular}{c|ccc|ccc|ccc}
    \Xhline{2\arrayrulewidth}
     & \multicolumn{3}{c|}{\gcop} & \multicolumn{3}{c|}{\yelp} & \multicolumn{3}{c}{\insr} \\
    \Xhline{\arrayrulewidth}
     & \multicolumn{3}{c|}{Victim Model} & \multicolumn{3}{c|}{Victim Model} & \multicolumn{3}{c}{Victim Model} \\
    Attack Method & \caregnn & \pcgnn & \gaga & \caregnn & \pcgnn & \gaga & \caregnn & \pcgnn & \gaga \\
    \Xhline{\arrayrulewidth}
    \clean & 48.02 & 55.62 & 21.68 & 29.79 & 59.13 & 28.00 & 16.42 & 16.17 & 15.68 \\
    \gnia    & \underline{60.67$\pm$0.21} & \underline{66.25$\pm$0.24} & \underline{25.76$\pm$0.32} & \underline{34.81$\pm$0.30} & \underline{63.57$\pm$0.17} & \underline{28.83$\pm$0.44} & 15.89$\pm$0.67 & 16.27$\pm$0.76 & 17.13$\pm$0.40 \\
    \tdgia   & 55.17$\pm$0.03 & 60.72$\pm$0.02 & 24.59$\pm$0.04 & 30.08$\pm$0.09 & 58.33$\pm$0.10 & 28.23$\pm$0.04 & \underline{18.34$\pm$0.03} & 18.25$\pm$0.04 & \underline{23.38$\pm$0.08} \\
    \cluster & 57.10$\pm$0.10 & 61.82$\pm$0.07 & 22.13$\pm$0.06 & 30.02$\pm$0.08 & 58.41$\pm$0.15 & 27.99$\pm$0.04 & N/A & N/A & N/A \\
    \gsquare & OOM & OOM & OOM & OOM & OOM & OOM & 17.24$\pm$0.00 & \textbf{20.08$\pm$0.04} & OOM \\
    \ours    & \textbf{88.40$\pm$0.42} & \textbf{89.36$\pm$0.69} & \textbf{41.34$\pm$1.90} & \textbf{55.59$\pm$2.94} & \textbf{94.21$\pm$1.79} & \textbf{29.63$\pm$0.54} & \textbf{18.63$\pm$0.63} & \underline{19.78$\pm$0.26} & \textbf{27.25$\pm$1.59} \\
    \Xhline{2\arrayrulewidth}
\end{tabular}

\caption{Multi-target attack performance on \gcop, \yelp, and \insr where \gcn is the surrogate model. We report misclassification rates (\%). OOM: Out of Memory. N/A: Not completed in 5 days.}
\label{tb:mainexp2}

\end{table*}
\paragraph{\normalfont{\textbf{Fraud Detectors, Baselines, and Implementation Details}}}
We consider following models as surrogate and victim models: \gcn~\cite{gcn}, \gsage~\cite{gsage}, \gat~\cite{gat}, \caregnn~\cite{caregnn}, \pcgnn~\cite{pcgnn}, and \gaga~\cite{gaga}. Since we focus on graph injection evasion attacks where attack nodes are injected into the original graph during the inference phase, we exclude fraud detectors that cannot handle the nodes added after training, such as BWGNN~\cite{bwgnn}, SplitGNN~\cite{split}, and DGA-GNN~\cite{dgagnn}. We use \gnia~\cite{gnia}, \tdgia~\cite{tdgia}, \cluster~\cite{cluster}, and \gsquare~\cite{g2a2c} as attack baselines. Further details on the experimental environment and implementation details of the fraud detectors and attack baselines are available in Appendix~\ref{sec:app_impl_baseline}, and implementation details of \ours are described in Appendix~\ref{sec:app_impl_ours}.

\subsection{Performance of Multi-target Attacks}
Table~\ref{tb:mainexp1} presents the results of multi-target attacks when the types of surrogate and victim models are the same. We repeat all experiments five times and report the average and standard deviation of the misclassification rates of all target sets weighted by their sizes, as the size varies across different target sets. \clean represents the misclassification rates on the original clean graphs. We see that GNN-based fraud detectors are more robust than vanilla GNNs due to their ability to handle heterophily. \ours shows the best attack performance across all datasets, as it can search adversarial structures more extensively than other methods.

Table~\ref{tb:mainexp2} shows the results in a more realistic and challenging scenario where \gcn is the surrogate model and \caregnn, \pcgnn, and \gaga are victim models. \ours outperforms all baselines, even when merely using \gcn as the surrogate model. This demonstrates that \ours effectively generalizes its attacks by comprehensively capturing and leveraging the interdependencies between node attributes and edges, as well as among target, candidate, and attack nodes. More detailed analyses are provided in Appendix~\ref{subsec:app_multi}.

\subsection{Ablation Studies and Efficiency Analysis of \ours}
Table~\ref{tb:ablation} shows the results of ablation studies on \gcop for \ours with the same settings as in Table~\ref{tb:mainexp2}. We replace the adversarial attribute and edge generation methods with random generation. For attributes, nodes are randomly selected from the original graph, and their attributes are assigned to attack nodes (random attributes). For edges, connections are randomly created among target, candidate, and attack nodes (random edges). We remove the learnable positional encodings $\vbp$, $\vtp$, and $\vhp$ (w/o pos. encoding) and modify \ours to select the candidates randomly (random candidates). Lastly, we fix the degree budget for each attack node (fixed budget). In conclusion, the original \ours shows the best performance, demonstrating the importance of each component of \ours for effective graph injection attacks. More ablation studies are presented in Appendix~\ref{subsec:app_ablation}.

We also compare \ours with attack baselines in terms of runtime and memory usage on \gcop, \yelp, and \insr, using \gcn as the surrogate model and \gaga as the victim model in Appendix~\ref{subsec:app_efficiency}. Overall, \ours is the most efficient in terms of runtime while maintaining moderate memory usage. This is because \ours effectively narrows the search space through candidate selection and adopts efficient matrix operations to inject all attack nodes at once. In addition, the complexity analysis of \ours is provided in Appendix~\ref{subsec:app_complexity}.

\subsection{Case Study: Effects of the Size of Fraud Gangs}
To analyze vulnerabilities of GNN-based fraud detectors regarding the size of fraud gangs, we categorize target sets in \gcop into three groups based on $B = \lvert \sN^{(1)} \cup \sT \rvert$, which reflects the size of the fraud gang. Table~\ref{tb:exp_case} presents the multi-target attack performance of \gnia and \ours for each category, using \gcn as the surrogate model. The results show that the disparity in misclassification rates before and after the attack widens with increasing the size of gangs ($B$). This highlights the threats that large-scale fraud gangs pose to GNNs. Specifically, \caregnn and \pcgnn, which only filter node-level camouflages, are more vulnerable to relatively large fraud gangs ($B > 10$) compared to \gaga. Notably, the performance gap between \gnia and \ours becomes larger as the size of gangs increases, suggesting that \ours can explore the complex structures that could be formed by large fraud gangs more comprehensively than \gnia.

\begin{figure}[t]
\centering
\includegraphics[width=0.97\columnwidth]{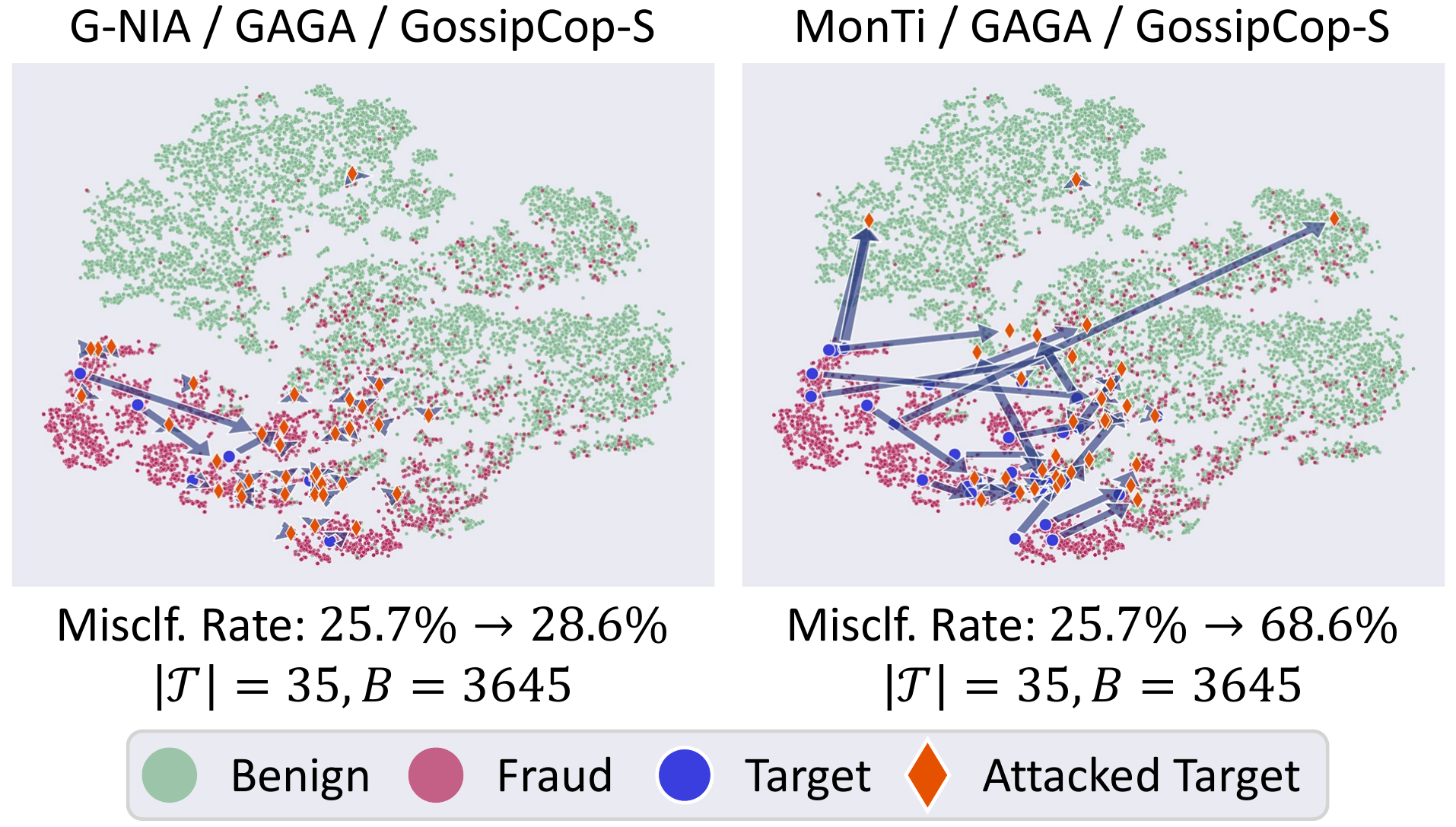}
\caption{The t-SNE visualization of the changes in the latent representations of target nodes computed by \gaga on \gcop, incurred by \gnia (Left) and \ours (Right).}
\label{fig:case}
\end{figure}

To analyze cases where \gnia fails but \ours succeeds, we show a case study on \gcop using target sets with $B > 1000$ and \gcn as the surrogate model. We visualize the latent node representations computed by \gaga before and after the attack using t-SNE~\cite{tsne} in Figure~\ref{fig:case} and Appendix~\ref{subsec:app_case}. Blue circles indicate the representations of target nodes before the attack, and orange diamonds indicate those after the attack. A blue circle and an orange diamond that correspond to the same target node are connected. Below each t-SNE visualization, we provide misclassification rates before and after the attack, the size of the target set, and $B$. \gnia results in minor changes in target node representations, leading to a small increase in misclassification rates. In contrast, \ours significantly shifts the representations from the fraud to the benign area, effectively making most target nodes misclassified.

\begin{table}[t]
\small
\renewcommand{\arraystretch}{1.05}
\setlength{\tabcolsep}{0.25em}
\centering

\begin{tabular}{c|C{5.2em}C{5.2em}C{5.2em}}
\Xhline{2\arrayrulewidth}
 & \caregnn & \pcgnn & \gaga \\ \hline
\clean & 48.02 & 55.62 & 21.68 \\
\Xhline{\arrayrulewidth}
random attributes & 78.18 & 77.81 & 25.70 \\
random edges & 79.64 & 81.40 & 39.33 \\
w/o pos. encoding & 86.12 & 86.98 & 40.21 \\
random candidates & 88.25 & 88.38 & 39.13 \\
fixed budget & 87.28 & 88.66 & 42.17 \\
\Xhline{\arrayrulewidth}
\ours & \textbf{88.78} & \textbf{89.90} & \textbf{43.70} \\
\Xhline{2\arrayrulewidth}
\end{tabular}

\caption{Ablation studies of \ours on \gcop with surrogate \gcn. We report misclassification rates (\%).}
\label{tb:ablation}

\end{table}

\begin{table}[t]
\small
\renewcommand{\arraystretch}{1.05}
\setlength{\tabcolsep}{0.5em}
\centering

\begin{tabular}{cC{5.5em}C{5.5em}C{5.5em}}
    \Xhline{2\arrayrulewidth}
    & \caregnn & \pcgnn & \gaga \\
    \Xhline{\arrayrulewidth}
    
    \multicolumn{4}{c}{$B \leq 10$ (\#Sets = 277)} \\
    \Xhline{\arrayrulewidth}
    
    \clean & 48.04\% & 54.46\% & 16.39\% \\
    \gnia & 49.17\% & 54.53\% & 17.75\% \\
    \ours & \textbf{58.69\%} & \textbf{61.63\%} & \textbf{27.95\%} \\

    \Xhline{\arrayrulewidth}
    \multicolumn{4}{c}{$10 < B \leq 1000$ (\#Sets = 316)} \\
    \Xhline{\arrayrulewidth}
    
    \clean & 51.30\% & 58.35\% & 11.59\% \\
    \gnia & 56.47\% & 63.73\% & 15.18\% \\
    \ours & \textbf{83.39\%} & \textbf{86.44\%} & \textbf{27.97\%} \\

    \Xhline{\arrayrulewidth}
    \multicolumn{4}{c}{$B > 1000$ (\#Sets = 316)} \\
    \Xhline{\arrayrulewidth}
    
    \clean & 55.00\% & 61.41\% & 28.66\%\\
    \gnia & 65.52\% & 70.45\% & 33.21\% \\
    \ours & \textbf{98.70\%} & \textbf{98.22\%} & \textbf{56.14\%} \\

    \Xhline{2\arrayrulewidth}
\end{tabular}

\caption{Multi-target attack performance on \gcop using \gcn as the surrogate model, with target sets categorized into three groups based on $B$. \#Sets denotes the number of target sets within each category.}
\label{tb:exp_case}

\end{table}

\section{Discussion on Defenses against \ours}

Our research aims to investigate the vulnerabilities of GNN-based fraud detectors and highlight their risks by simulating attacks of fraud gangs in practical settings. In particular, our findings suggest that GNN-based fraud detectors are especially susceptible to large-scale group attacks, which could undermine their reliability in real-world applications. We discuss potential approaches that could be beneficial to safeguarding against such threats. Our promising approach is to extend existing graph adversarial defense methods~\cite{gnnguard, mmd}, such that those methods can be applied to GNN-based fraud detectors by identifying collusive patterns of fraudulent nodes. For instance, rather than solely focusing on individual nodes to detect malicious behavior, considering how the neighborhood structure around target nodes is partitioned into distinct groups could provide more effective defenses against attacks of fraud gangs. Additionally, incorporating adversarial training~\cite{survey1, survey2} with generative models could also be helpful in developing community-aware GNN-based fraud detectors. We believe that these approaches can mitigate the impact of gang-level attacks and improve the robustness of GNN-based fraud detection methods.

\section{Conclusion and Future Work}
We investigate adversarial attacks against GNN-based fraud detectors with various practical scenarios and real-world datasets. To the best of our knowledge, our work is the first study to explore such attacks against GNN-based fraud detectors and graph injection attacks for multiple target nodes formed by fraud gangs. We define this task as a multi-target graph injection attack and propose \ours, a new adversarial attack model that injects all attack nodes at once. \ours employs adaptive degree budget allocation for each node to explore diverse injection structures. Furthermore, \ours simultaneously generates attributes and edges of attack nodes, considering the interdependency between them. Extensive experiments on five real-world graphs show that \ours outperforms state-of-the-art graph injection attack methods in both multi- and single-target settings. For future work, we will extend \ours to multi-relational graphs~\cite{drag}, as considering multiple relation types can be beneficial to effective fraud detection. We also plan to examine our work in inductive settings~\cite{ingram} where both victim and attack models do not observe test nodes and their connections during training. Moreover, we will theoretically investigate \ours in terms of generalization bounds~\cite{reed}.

\section*{Acknowledgments}
This research was supported by an NRF grant funded by MSIT 2022R1A2C4001594 (Extendable Graph Representation Learning) and an IITP grant funded by MSIT 2022-0-00369, RS-2022-II220369 (Development of AI Technology to support Expert Decision-making that can Explain the Reasons/Grounds for Judgment Results based on Expert Knowledge).

\bibliography{aaai25}

\appendix
\setcounter{secnumdepth}{2} 
\onecolumn
\section*{Appendix}
\section{Detailed Discussions on Related Work}
\label{sec:app_rel_work}
\subsection{Graph-based Fraud Detection}
Graph-based fraud detection aims to identify entities associated with illicit activities by considering graph structures. It has been widely applied across various practical scenarios, such as detecting fake news~\cite{gossip, upfd, get}, identifying spam reviews~\cite{caregnn, pcgnn}, and uncovering insurance frauds~\cite{rdpgl}. Recently proposed fraud detection methods have adopted GNNs to detect malicious activities by aggregating the neighborhood information~\cite{caregnn, pcgnn, h2fd, dignn, dagnn, gaga}. 
These GNN-based fruad detection methods define this task as a problem of classifying nodes into two categories: fraud or benign.

In fraud graphs, fraud nodes are much fewer than benign nodes, resulting in a class imbalance problem. In addition, fraud nodes tend to connect with benign nodes or imitate attributes of benign nodes to conceal their malicious activities. Such behavior introduces heterophily~\cite{hetero}, which makes vanilla GNNs challenging to distinguish between fraud and benign nodes. The class imbalance problem and heterophily in graphs have been raised as significant challenges in GNN-based fraud detection. CARE-GNN~\cite{caregnn} defines this phenomenon as camouflage and addresses it by employing a neighborhood selector with reinforcement learning. FRAUDRE~\cite{fraudre} introduces an imbalance-oriented classification module to address the training bias resulting from the large number of benign nodes. PC-GNN~\cite{pcgnn} devises a neighborhood sampler based on a learnable distance function, and a label-balanced graph sampler to choose neighbors for aggregation. BWGNN~\cite{bwgnn} and SplitGNN~\cite{split} alleviate the heterophily utilizing the Beta Wavelet in the lens of the spectral domain. Several methods have also been proposed to explicitly distinguish and separately leverage heterophilic and homophilic edges~\cite{h2fd, dignn, dagnn}. GAGA~\cite{gaga} proposes the group aggregation strategy that leverages observable labels, neighbor distances, and relationships, using a transformer encoder. DGA-GNN~\cite{dgagnn} addresses the non-additivity of node features by proposing decision tree binning encoding and introduces a dynamic grouping strategy to improve group distinguishability.

Despite the recent advancements, vulnerabilities in these GNN-based fraud detectors have not yet been studied. In this work, we investigate adversarial attacks against GNN-based fraud detectors in real-world scenarios. We focus on graph injection evasion attacks where attack nodes are injected into the original graph during the inference phase. Thus, we have excluded the following fraud detection methods from our experiments, as their inability to handle the nodes added after training hinders a fair comparison with other GNN-based fraud detectors:  BWGNN~\cite{bwgnn} and SplitGNN~\cite{split} employ graph spectral filters that rely on the eigenvectors and eigenvalues of the graph laplacian matrix, which limits their ability to handle new nodes added after training. DGA-GNN~\cite{dgagnn} introduces a dynamic grouping mechanism to enhance the distinguishability of grouped messages. However, since this mechanism groups nodes based on predictions generated during training, it cannot handle new nodes added after training.

\subsection{Adversarial Attacks on Graphs}

\subsubsection{Taxnomies of Adversarial Attacks on Graphs: Attack Stage and Attackers' Knowledge}
It has been recently highlighted that vanilla GNNs are vulnerable to adversarial attacks~\cite{gp, rewatt, spac, infmax}. Adversarial attacks on graphs can be carried out in various settings depending on the problem setting~\cite{survey1, survey2, marl}. In terms of the attack stage, the attack can be categorized as either a poisoning attack or an evasion attack. A poisoning attack injects malicious data into the training set, which causes the victim model to malfunction~\cite{survey1, survey2}. However, given the massive scale and accessibility of training data, the poisoning attack has been considered as highly cost-inefficient and challenging~\cite{marl}. On the other hand, an evasion attack introduces defects into the victim model solely during the inference phase, which makes it more realistic for attacks against GNN-based fraud detectors. 

The attack can be also classified as either a white-box or black-box attack, depending on the attacker's knowledge. In a white-box attack, it is assumed that the attacker can access all information about the architecture and parameters of the victim model~\cite{survey2}. However, acquiring such details about the victim model is practically challenging. In contrast, a black-box attack is more feasible since it supposes that the attacker cannot directly access the victim model~\cite{tdgia}. To conduct attacks in the most realistic setting, we adopt a black-box evasion attack setting for our study of multi-target graph injection attacks against GNN-based fraud detectors.

\subsubsection{Graph Modification Attacks and Graph Injection Attacks}
It has been shown that even a few adversarial modifications on graphs can significantly degrade the performance of GNNs~\cite{nettack, metattack, rewatt, infmax, spac}. For instance, Nettack~\cite{nettack} generates adversarial perturbations on node attributes and edges while preserving the node degree distribution. Meta-attack~\cite{metattack} modifies the graph structure using meta-gradients, under constraints of the unnoticeability. Graph injection attacks have emerged focusing on more practical settings, where attackers inject new malicious nodes instead of modifying existing nodes and edges~\cite{afgsm, nipa}. NIPA~\cite{nipa} employs reinforcement learning to inject malicious nodes, while AFGSM~\cite{afgsm} adapts the fast gradient sign method to efficiently inject nodes. However, although both NIPA and AFGSM involve injection attacks, their poisoning attack setting requires retraining the victim model on the attacked graph, which might be demanding in real-world scenarios.

Meanwhile, graph injection evasion attack methods have recently been proposed, offering a more feasible alternative to graph poisoning attacks~\cite{gnia, tdgia, cluster, g2a2c}. For example, \gnia~\cite{gnia} jointly learns attribute and edge generators to attack unseen target nodes, whereas \tdgia~\cite{tdgia} determines where to connect attack nodes using the defective edge selection strategy and then optimizes the features of the injected nodes. \cluster~\cite{cluster} regards graph injection attack as a clustering task, clustering target nodes by adversarial vulnerability, assigning attack nodes to each cluster, and optimizing their attributes. \gsquare~\cite{g2a2c} aims to alleviate the discrepancies between surrogate and victim models, based on the advantage actor-critic framework. In our experiments, we have employed attack methods that align with the multi-target graph injection attack under a black-box evasion attack setting, such as \gnia, \tdgia, \cluster, and \gsquare.

\subsubsection{Comparison between Existing Graph Injection Attack Methods and \ours}
Despite the recent progress in graph injection attack methods, they still have several limitations when applied to multi-target attacks against GNN-based fraud detectors. First, the existing methods, such as \gnia~\cite{gnia}, \tdgia~\cite{tdgia}, \cluster~\cite{cluster}, and \gsquare~\cite{g2a2c}, assign the fixed degree budget to each attack node or inject attack nodes sequentially. This limits their flexibility and efficiency in exploring diverse structures, as it requires to fix the degree budget across all attack nodes due to a lack of information about future steps. Second, the existing methods focus on single-target~\cite{g2a2c} or single-injection attacks~\cite{gnia}. As a result, they often overlook interactions within target nodes and among attack nodes, which are crucial for modeling the intricate relationships within a fraud gang and attacking target nodes. Third, the existing methods sequentially generate attributes and edges of attack nodes, considering a one-way dependency, which fails to accurately model collusive patterns of fraud gangs. In contrast, \ours addresses these limitations by 1) injecting all attack nodes at once, adaptively assigning the degree budget for each attack node to explore diverse structures among target, candidate, and attack nodes, 2) employing a transformer encoder to effectively capture interdependencies within target and attack nodes, and 3) simultaneously generating attributes and edges of attack nodes to capture their interdependency.

\section{Details about Datasets for Multi-target Attacks}
\label{sec:app_data}
We have created datasets and target sets consisting of fraud nodes which are grouped based on their metadata or relations in real-world graphs: \gcop, \yelp, and \insr. Detailed descriptions of each graph and the process of generating the target sets are provided below, and statistic of the three datasets is presented in Table~\ref{tb:app_data}. Notably, our datasets are substantially larger and denser than those typically used in the previous graph adversarial attack studies~\cite{gnia, tdgia, cluster, g2a2c}. Our datasets include \gcop (16,488 nodes, 3,865,058 edges), \yelp (45,900 nodes, 3,846,910 edges), and \insr (122,792 nodes, 912,833 edges). In contrast, among the datasets used in our attack baselines, the largest is \pubmed~\cite{pubmed} (19,717 nodes, 44,338 edges), and the densest is \wikics~\cite{wikics} (11,701 nodes, 216,123 edges). In experiments, we extract the largest connected component of each dataset. We set the percentage of training nodes $p$, the parameter to control node budgets $\rho$, and the parameter to control edge budgets $\xi$ for each dataset based on its characteristics to ensure that the attacks are performed under realistic and challenging conditions. Further details on $\rho$ and $\xi$ are provided in Appendix~\ref{subsec:app_budget}.

\begin{table}[t]
\renewcommand{\arraystretch}{1.2}
\setlength{\tabcolsep}{0.5em}
\centering

\begin{tabular}{c|C{5.5em}C{5.5em}C{5.5em}}
    \Xhline{2\arrayrulewidth}
    & \gcop & \yelp & \insr \\
    \Xhline{\arrayrulewidth}
    The type of node features & Continuous & Continuous & Discrete \\
    $|\sV|$ & 16,488 & 45,900 & 122,792 \\
    The number of frauds & 3,898 & 6,656 & 1,264 \\
    The number of target sets & 2,438 & 1,435 & 380 \\
    $\overline{|\sT_{\text{train}}|}$ & 11.96 & 4.82 & 3.14 \\
    $\overline{|\sT_{\text{val}}|}$ & 8.38 & 3.56 & 4.02 \\
    $\overline{|\sT_{\text{test}}|}$ & 11.57 & 4.89 & 5.51 \\
    $|\sE|$ & 3,865,058 & 3,846,910 & 912,833 \\
    $D$ & 768 & 32 & 1,611 \\
    $\Delta$ & 52.00 & 25.39 & 18.64 \\
    $\eta$ & 2355.70 & 2119.56 & 130.09 \\
    \Xhline{2\arrayrulewidth}
\end{tabular}

\caption{Statistic of datasets for multi-target attacks. $\overline{|\sT_{\text{train}}|}$, $\overline{|\sT_{\text{val}}|}$, and $\overline{|\sT_{\text{test}}|}$ represent the average sizes of the train, validation, and test target sets, respectively. For $\Delta \, \text{and} \; \eta$, we report mean values.}
\label{tb:app_data}

\end{table}

\paragraph{\normalfont{\textbf{\gcop}}}
The original \gcoporg dataset contains news articles and the social user engagements of the news articles from Twitter~\cite{fnewsnet}. We generate \gcop by utilizing users' social engagements with the news articles to create an undirected weighted graph by referring to~\cite{gossip}. In \gcop, nodes represent news articles and edges are created when multiple users tweet two news articles. Note that we retain only the edges with the top 5\% engagements to prevent the graph from being excessively dense. While removing edges, we preserve the minimum spanning tree to maintain the connectivity of the graph. Fake news articles are defined as fraud nodes. When fake news articles are tweeted by the same user, we consider this as collusive behavior to spread misinformation. Thus, we group the fake news articles tweeted by the same user into a target set.

\paragraph{\normalfont{\textbf{\yelp}}}
The \yelp dataset ~\cite{yelp, fake} is a review graph where nodes represent reviews, and edges are created based on three different types of relations: (1) the reviews are written by the same user, (2) the reviews have the same star rating for the same product, and (3) the reviews are written within the same month for the same product. In \yelp, fake reviews are defined as fraud nodes. We consider fake reviews written by the same user, or fake reviews for the same product within the same month as evidence of collusive fraudulent behaviors. Thus, we group the fake reviews of the same user or the fake reviews for the same product within the same month into a target set. 

\paragraph{\normalfont{\textbf{\insr}}}
The \insr dataset is a medical insurance graph generated based on claim data collected by an anonymous insurance company. In \insr, nodes correspond to claims, and edges represent relationships predefined by domain experts. We group fraud claims into a target set, guided by domain experts of the insurance company that provided the data. Due to the request from the company, we are unable to provide further details.

\section{More Details about Fraud Detectors and Attack Baselines}
All experiments, except those on \insr, are conducted using GeForce RTX 2080 Ti, RTX 3090, or RTX A6000 with Intel(R) Xeon(R) Gold 6330 CPU @ 2.00GHz, running on Ubuntu 18.04.5 LTS. Experiments on \insr are performed on an Amazon EC2 G4dn.metal instance with eight Nvidia T4 GPUs and 384GB RAM. Implementation details of the fraud detectors and attack baselines are described below.
\label{sec:app_impl_baseline}

\subsection{Details about Surrogate and Victim Fraud Detection Models}
We consider following models as surrogate and victim models: \gcn~\cite{gcn}, \gsage~\cite{gsage}, \gat~\cite{gat}, \caregnn~\cite{caregnn}, \pcgnn~\cite{pcgnn}, and \gaga~\cite{gaga}. We train all the methods with two different initialization seeds utilizing the Adam optimizer~\cite{adam} by following the strategy used in ~\cite{tdgia}; the models initialized with the first seed serve as surrogate models, while those initialized with the second seed are employed as victim models. The weighted cross entropy is employed as the loss function to address the class imbalance issue. We conduct validation every 10 epochs up to a maximum of 1000 epochs, applying early stopping with the patience of 100. For all methods, we set the hidden dimension to 64 and the batch size to 1,024. The performance of victim models on clean graphs, measured in terms of \fone and \auc, is reported in Table~\ref{tb:app_clean}. \fone provides a balanced measure of precision and recall for all classes, while \auc evaluates the model's ability to distinguish between fraud and benign classes. For both \fone and \auc, higher values indicate better performance.

\begin{table}[t]
\renewcommand{\arraystretch}{1.2}
\setlength{\tabcolsep}{0.5em}

\centering

\begin{tabular}{c|C{4em}C{4em}|C{4em}C{4em}|C{4em}C{4em}}
    \Xhline{2\arrayrulewidth}
    & \multicolumn{2}{c|}{\gcop} & \multicolumn{2}{c|}{\yelp} & \multicolumn{2}{c}{\insr} \\
    & \fone & \auc & \fone & \auc & \fone & \auc \\
    \Xhline{\arrayrulewidth}
    \gcn & 0.8349 & 0.9250 & 0.5481 & 0.5798 & 0.8218 & 0.9045 \\
    \gsage & 0.8456 & 0.9336 & 0.7173 & 0.8463 & 0.8963 & 0.9314 \\
    \gat & 0.8794 & 0.9540 & 0.6654 & 0.7977 & 0.8155 & 0.9055 \\
    \caregnn & 0.7514 & 0.8637 & 0.6619 & 0.8240 & 0.8645 & 0.9261 \\
    \pcgnn & 0.6820 & 0.7794 & 0.6537 & 0.7688 & 0.8650 & 0.9249 \\
    \gaga & 0.8803 & 0.9570 & 0.7470 & 0.8962 & 0.8845 & 0.9313 \\
    \Xhline{2\arrayrulewidth}
\end{tabular}

\caption{Fraud detection performance of victim models on clean \gcop, \yelp, and \insr. We report \fone and \auc. For both \fone and \auc, higher values indicate better performance.}
\label{tb:app_clean}

\end{table}

The general GNNs, \gcn, \gsage, and \gat, are implemented with two layers, followed by a 2-layer MLP classifier, using \dgl~\cite{dgl} and \torch~\cite{pytorch}. To find the optimal hyperparameters for these methods, we perform a grid search, exploring learning rates in $\{0.01, 0.001\}$ and weight decay values in $\{0.001, 0.0001\}$. The mean aggregator is adopted as the aggregation function of \gsage. For \gat, we utilize multi-head attention and tuned the number of attention heads in $\{2, 8\}$.

For \caregnn, \pcgnn, and \gaga, we set the hyperparameters of them by following the official codes or descriptions provided in the methods' original publications. Additionally, the original implementation of \gaga for generating and updating group sequences of nodes does not utilize efficient operations like sparse matrix or tensor operations, leading to substantial computational overheads. Therefore, based on sparse matrix and tensor operations, we implemente an efficient method to update group sequences of the attacked target nodes, strictly maintaining the functionality of the original implementation.

\subsection{Details about Attack Baselines}
We use the following methods as baselines for graph injection attack methods: \gnia~\cite{gnia}, \tdgia~\cite{tdgia}, \cluster, and \gsquare~\cite{g2a2c}. We set the hyperparameters of the baselines, including the maximum number of epochs and patience settings, based on the official code or as outlined in the original publications of the methods.

\paragraph{\normalfont{\textbf{\gnia \cite{gnia}}}} 
\gnia is originally designed to inject a single attack node. Therefore, we extend \gnia to enable the injection of multiple attack nodes, making it suitable for multi-target graph injection attacks. Meanwhile, as the sequential injection approach causes out-of-memory issues, we adopt the one-time injection approach, which concatenates random noise vectors from the standard normal distribution to the input of the attribute generation module.

\paragraph{\normalfont{\textbf{\tdgia \cite{tdgia}}}}
Since the official implementation of \tdgia only supports attacks on datasets with continuous node attributes, we extend \tdgia by introducing the discrete feature optimization through the top-$k$ operation and straight-through estimator~\cite{straight} to enable attacks on \insr dataset whose node attributes are discrete.

\paragraph{\normalfont{\textbf{\cluster \cite{cluster}}}}
\cluster learns to inject attack nodes by directly querying the victim model. However, since access to the victim model is not feasible in our scenario, we instead adopt an alternative version of \cluster that utilizes gradients from a surrogate model, as implemented in the official code.

\paragraph{\normalfont{\textbf{\gsquare \cite{g2a2c}}}}
The original \gsquare is only capable of single-target attacks. Therefore, we extend \gsquare to enable multi-target attacks. Following \gnia~\cite{gnia}, we employ the mean of representations of target nodes as a representation of the target set and utilize it as the input for the adversarial node generator. For the adversarial edge sampler and the value predictor, we adjust them to consider all nodes in the target set. Extra rewards, which are originally given for each successful attack on a single target node, are weighted by the overall attack success rate across all target nodes in the target set.

\section{Implementation Details of \ours}
\label{sec:app_impl_ours}
\ours is implemented using \dgl~\cite{dgl} and \torch~\cite{pytorch}. To identify the optimal hyperparameters of \ours, we perform a grid search, which includes whether \ours adopts the attack node masking strategy or not. Our search covers learning rates in $\{0.01, 0.001\}$, weight decay in $\{0.001, 0.0001\}$, dropout rates in $\{0, 0.1, 0.2\}$, and dimensions of the feed-forward network of the adversarial structure encoding transformer in $\{512, 2048\}$. Key parameters of \ours are set as follows: the number of neighbor hops for candidate node selection $K=2$, the maximum number of candidates $n_c=128$, the number of layers $L=6$, the number of heads $n_h=4$, and the hidden dimension $D_H=64$. We apply exponential decay to temperature $\tau$ and exploration parameter $\epsilon$, reducing them from 10 to 0.01 at a decay rate of 0.63 per epoch. \ours is trained using the Adam optimizer~\cite{adam} over $100$ epochs, applying early stopping with a patience of $10$. In the case of single node injection, where the node budget $\Delta=1$, the representation of the attack node $\vhz_1$ is initialized to the zero vector. For single-target attack experiments, which utilize multi-class node classification datasets, we optimize \ours using the C\&W loss~\cite{loss, gnia, cluster} which is the multi-class version of our loss function.

\section{Complete Experimental Results and Additional Analyses}
This section presents more experimental results and analyses, expanding those summarized or mentioned in the main paper.

\subsection{Results of Single-target Attacks}
\label{subsec:app_single}
Although \ours is designed for multi-target attacks, we present benchmark results for single-target attacks on \ogbp~\cite{ogbp} and \pubmed~\cite{pubmed} using \gcn as the surrogate and victim models to verify that \ours is also applicable in general scenarios. The \ogbp is a co-purchasing network of Amazon products. The \pubmed is a citation network from the PubMed database. For the single-target attack experiments, we set the node budget $\Delta=1$ and edge budget $\eta=1$, following~\cite{gnia, g2a2c}. In this setting, we additionally consider the following baselines: \random~\cite{gnia}, \nettack~\cite{nettack} with a randomly initialized attack node (denoted by Rand+\nettack), \nipa~\cite{nipa}, and \afgsm~\cite{afgsm}. Following the experimental setup of ~\cite{g2a2c}, we use the subgraph and data splits distributed by \gnia~\cite{gnia} for \ogbp, and public splits for \pubmed. Results are presented in Table~\ref{tb:app_benchmark}. \clean represents the misclassification rates of the victim \gcn measured on the original clean graph. The results labeled as “(obtained from~\cite{g2a2c})”, such as Rand+\nettack, \nipa, and \afgsm, are obtained from~\cite{g2a2c}, whereas the other results (i.e., \clean, \random, \gnia, \tdgia, \gsquare, and \ours) were obtained by running the methods by ourselves. For the results obtained from~\cite{g2a2c}, we also present \clean (obtained from~\cite{g2a2c}), which denotes the misclassification rates of the victim \gcn on the original clean graph that have been reported in~\cite{g2a2c}. Despite not being specifically designed for single-target attacks, \ours achieves the highest misclassification rates on both datasets. This demonstrates the effectiveness of \ours's transformer-based architecture, which captures intricate interactions around a target node via adversarial structure encoding.

\begin{table}[t]
\renewcommand{\arraystretch}{1.2}
\setlength{\tabcolsep}{1.0em}

\centering

\begin{tabular}{c|C{4.5em}C{4.5em}}
\Xhline{2\arrayrulewidth}
 & \ogbp & \pubmed \\
\Xhline{\arrayrulewidth}
\clean & 20.3 & 24.3 \\
\clean (obtained from~\cite{g2a2c}) & 24.3 & 21.9 \\
\Xhline{\arrayrulewidth}
\random & 25.0$\pm$0.5 & 24.9$\pm$0.6 \\
Rand+\nettack (obtained from~\cite{g2a2c}) & 63.3$\pm$0.5 & 46.7$\pm$0.6 \\
\nipa (obtained from~\cite{g2a2c}) & 25.9$\pm$0.2 & 21.9$\pm$0.0 \\
\afgsm (obtained from~\cite{g2a2c}) & 74.9$\pm$0.7 & 65.8$\pm$0.9 \\
\gnia & \underline{94.9$\pm$0.2} & 68.0$\pm$1.8 \\
\tdgia & 83.8$\pm$0.6 & 47.2$\pm$0.9 \\
\gsquare & 92.3$\pm$0.5 & \underline{82.0$\pm$0.0} \\
\ours & \textbf{96.4$\pm$1.1} & \textbf{85.3$\pm$1.1} \\
\Xhline{2\arrayrulewidth}
\end{tabular}

\caption{Single-target attack performance on \ogbp and \pubmed with \gcn as the surrogate and victim models. We report misclassification rates (\%). \clean represents the misclassification rates of the victim \gcn measured on the original clean graph. The results labeled as “(obtained from~\cite{g2a2c})”, such as Rand+\nettack, \nipa, and \afgsm, are obtained from~\cite{g2a2c}, whereas the other results (i.e., \clean, \random, \gnia, \tdgia, \gsquare, and \ours) were obtained by running the methods ourselves. For the results obtained from~\cite{g2a2c}, we also present \clean (obtained from~\cite{g2a2c}), which denotes the misclassification rates of the victim \gcn on the original clean graph that have been reported in~\cite{g2a2c}.}
\label{tb:app_benchmark}

\end{table}

\subsection{Budget Constraints for Multi-target Attacks: Formulation and Implications}
\label{subsec:app_budget}
\paragraph{\normalfont{\textbf{Rationales behind the Formulation of Budget Constraints}}}
This section explains the rationale for determining the total budget for a target set. Previous studies have determined budget constraints for a target set based on the degree of each target node~\cite{gnia} or by setting the fixed budget constraints with constant values~\cite{tdgia, cluster, g2a2c}. For attacks of fraud gangs, we can consider that the degrees of target nodes in a target set reflect the size and amount of available resources of the corresponding gang. Thus, the budget constraints for each target set have to be proportional to the overall degrees of the nodes in the target set. However, in multi-target attack scenarios where target nodes often share neighboring nodes, relying solely on the sum or average degree of the target nodes can make the budgets too large due to overlapping neighbors. To address this issue, we define the budgets based on the size of the one-hop neighbor set for each target set $B = \lvert \sN^{(1)} \cup \sT \rvert$. In addition, we also impose limits on the budgets since excessively large budgets can lead to highly noticeable and easy attacks. For each target set, the node budget $\Delta$ is defined to be proportional to the number of direct neighbors and constrained not to exceed the average level:
\begin{equation}
    \Delta=\max( \big\lfloor \rho \,\cdot\, \min(B, \, \overline{B} \,) + 0.5 \big\rfloor, \, 1)
\end{equation}
where $\rho$ is a parameter to control node budgets and $\overline{B}$ is the average value of $B$ across all target sets. The edge budget $\eta$ for each target set is determined based on both $\Delta$ and the average node degree of the target set, ensuring that the average node degree of the attack nodes remains below a predefined threshold relative to the average node degree of the entire graph:
\begin{equation}
    \eta=\Delta \,\cdot\, \max(\big\lfloor \min(d_\sT,\; \xi\cdot\overline{d} \,) + 0.5 \,\big\rfloor, \, 1)    
\end{equation}
where $d_\sT$ is the average node degree of the target set, $\xi$ is a parameter to control edge budgets, and $\overline{d}$ is the average node degree of all nodes in the graph. In our main experiments for multi-target attacks, we have determined the budget constraints, $\Delta$ and $\eta$, by setting $\rho$ and $\xi$ for each dataset considering its characteristics, and applied the same budget constraints for all attack methods. We set $\rho=0.05, \, \xi=0.1$ for \gcop, $\rho=0.05, \, \xi=0.5$ for \yelp, and $\rho=0.2, \, \xi=0.5$ for \insr. 

\begin{figure}[t]
\centering
\begin{subfigure}[h]{0.40\columnwidth}
    \centering
    \includegraphics[width=\columnwidth]{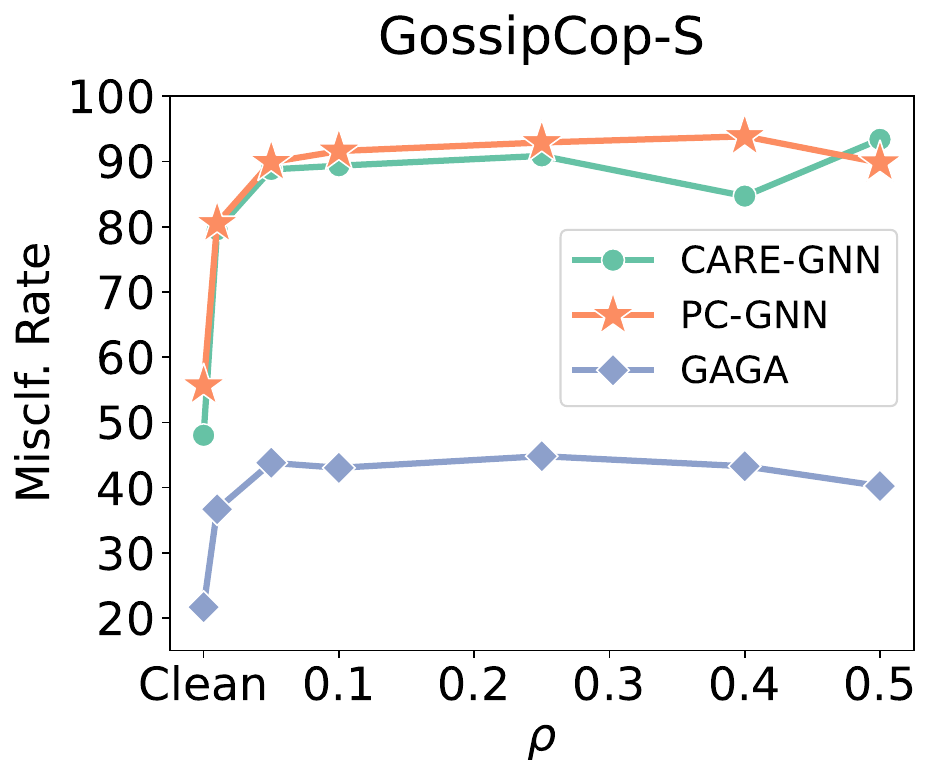}
\end{subfigure}
\begin{subfigure}[h]{0.349\columnwidth}
    \centering
    \includegraphics[width=\columnwidth]{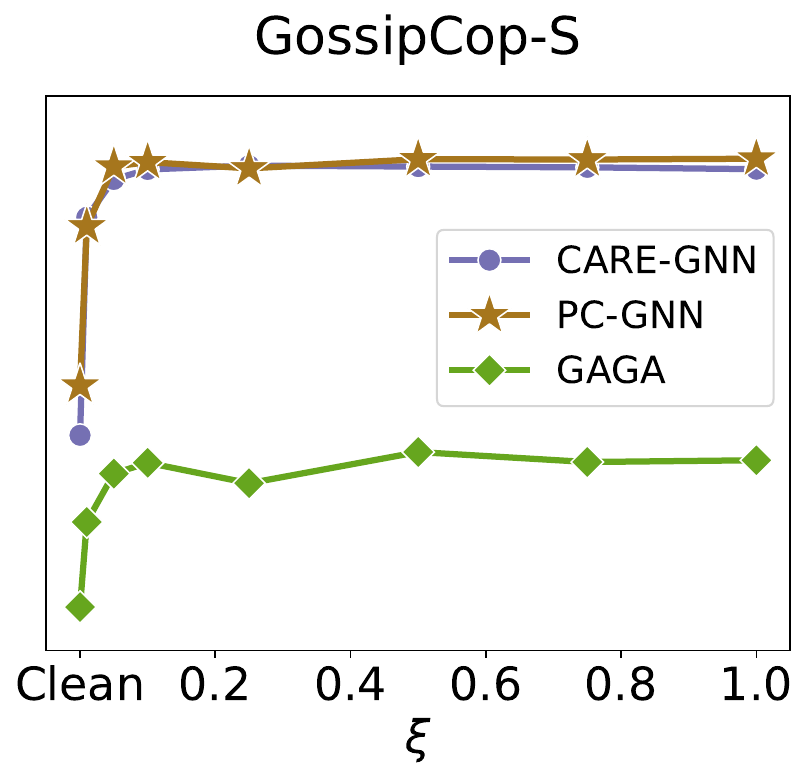}
\end{subfigure}
\caption{Multi-target attack performance of \ours on \gcop using \gcn as the surrogate model with varying node (Left) and edge budgets (Right).}
\label{fig:app_budget}
\end{figure}

\paragraph{\normalfont{\textbf{Analysis of the Effects of Varying $\rho$ and $\xi$}}}
To examine the influence and tendency of different budget constraints, we conduct additional experiments by varying the values of parameters to control budgets: $\rho$ and $\xi$. Figure~\ref{fig:app_budget} shows the attack performance of \ours under varying budgets on \gcop with \gcn as the surrogate model and \caregnn, \pcgnn, and \gaga as the victim models. For all cases, we see that misclassification rates initially increase sharply with a rise in budget, then seem to converge without significant further changes. 

\subsection{Results of Multi-target Attacks}
\label{subsec:app_multi}
The results of the multi-target attack experiments on \gcop, \yelp, and \insr are shown in Tables~\ref{tb:mainexp1} and~\ref{tb:mainexp2}. We see that GNN-based fraud detectors are generally more robust against attacks than vanilla GNNs due to their ability to handle heterophily. For example, CARE-GNN and PC-GNN employ a similarity-based neighborhood selection, which can be beneficial in alleviating the influence of injected attack nodes. Among the GNN-based fraud detectors, GAGA is relatively robust in general; GAGA alleviates the impacts of attack nodes through group aggregation by treating the attack nodes as unlabeled nodes.

\ours shows the best attack performance among all attack methods across all datasets since \ours can more extensively search for adversarial structures, including connections among attack nodes. We conduct independent samples t-tests with the significance level of 0.05 comparing \ours with the best-performing baselines for the results reported in Tables~\ref{tb:mainexp1} and~\ref{tb:mainexp2}. \ours outperforms all baselines in 25 cases, with statistically significant differences (with p-values less than 0.05) in 21 cases.  The performance variation of \ours across datasets can be attributed to their distinct characteristics. For example, \gcop has a denser structure and larger fraud gangs than the other graphs. This allows attackers to explore diverse injection structures, leading to more effective attacks. Also, the datasets differ in their node feature characteristics. For example, \gcop and \yelp use continuous features, which allow attackers to flexibly generate node attributes. In contrast, \insr contains discrete features that inherently limit the possible values for attack node features, posing challenges for all attack methods. While \ours's performance varies across datasets, \ours consistently outperforms all baselines. Furthermore, we evaluate \ours's impacts on non-target nodes across all datasets. Table~\ref{tb:app_nontarget} presents the misclassification rates of non-target test nodes before and after attacks by \ours, using \gcn as the surrogate model and \caregnn, \pcgnn, and \gaga as victim models. The results demonstrate that \ours's attacks have negligible impacts on the classification of non-target nodes. This is because \ours injects only a limited number of nodes around a target set, resulting in minimal impacts on non-target nodes.

Among the attack baseline methods, \gnia exhibits relatively superior performance among the graph injection attack baselines since \gnia can process the target nodes in the same target set together, unlike \tdgia and \cluster. However, \gnia oversimplifies interactions within target nodes by just using the mean representation of the target nodes and does not consider possible connections among attack nodes, resulting in significantly lower misclassification rates than \ours. Meanwhile, \tdgia and \cluster appear to encounter challenges in performing attacks. Since the defective edge selection strategy of \tdgia overlooks that edges can be formed among attack nodes, \tdgia struggles to consider appropriate connections among attack nodes in a sequential injection process. \cluster faces a challenge in specifying the proper number of clusters for target sets with varying sizes. In addition, for discrete features, \cluster applies perturbations to each element of features and computes the loss. As a result, on the \insr dataset, which has numerous feature dimensions and nodes, the computation time exceeds 5 days (N/A). In most settings, \gsquare encounters Out of Memory (OOM) errors, highlighting its limitations in scalability, which come from storing complete episodes of sequential injections.

\begin{table}[t]
\renewcommand{\arraystretch}{1.2}
\setlength{\tabcolsep}{0.5em}
\centering

\begin{tabular}{c|cc|cc|cc}
\Xhline{2\arrayrulewidth}
 & \multicolumn{2}{c|}{\gcop} & \multicolumn{2}{c|}{\yelp} & \multicolumn{2}{c}{\insr} \\
 & Before Attack & After Attack & Before Attack & After Attack & Before Attack & After Attack \\
\Xhline{\arrayrulewidth}
\caregnn & 18.72 & 18.65 $\pm$ 0.08 & 22.89 & 22.86 $\pm$ 0.02 & 0.52 & 0.52 $\pm$ 0.00 \\
\pcgnn & 24.54 & 24.48 $\pm$ 0.07 & 17.16 & 17.13 $\pm$ 0.02 & 0.52 & 0.52 $\pm$ 0.00 \\
\gaga & 8.76 & 8.73 $\pm$ 0.06 & 15.19 & 15.18 $\pm$ 0.01 & 0.42 & 0.42 $\pm$ 0.00 \\
\Xhline{2\arrayrulewidth}
\end{tabular}

\caption{Misclassification rates (\%) of non-target test nodes before and after attacks by \ours on \gcop, \yelp, and \insr using \gcn as the surrogate model and \caregnn, \pcgnn, and \gaga as victim models.}
\label{tb:app_nontarget}

\end{table}

\subsection{Additional Ablation Studies of \ours}
\label{subsec:app_ablation}
Table~\ref{tb:app_ablation} shows the results of ablation studies for \ours on \gcop, utilizing \caregnn, \pcgnn, and \gaga as victim models and \gcn as the surrogate model. We replace the adversarial attribute and edge generation methods with random generation. For attributes, nodes are randomly selected from the original graph, and their attributes are assigned to attack nodes (random attributes). For edges, connections are randomly created among target, candidate, and attack nodes (random edges). The attack performance of \ours significantly degrades when the random generation strategy is used. We also remove the learnable positional encodings $\vbp$, $\vtp$, and $\vhp$ (w/o pos. encoding) and degree information (w/o degree). Furthermore, we set the parameters to be shared between target and candidate encodings (shared parameters). In addition, we modify \ours to exclude candidate nodes during the entire process (w/o candidates) and to select the candidates randomly (random candidates). Lastly, we fix the degree budget for each attack node (fixed budget). While not as substantial as observed when replacing attribute and edge generation with the random strategy, there is a noticeable decline in performance when removing or modifying any of these modules. Overall, the original \ours shows the best attack performance, demonstrating the importance of each component of \ours in achieving effective graph injection attacks.

\begin{table}[t]
\renewcommand{\arraystretch}{1.2}
\setlength{\tabcolsep}{0.5em}
\centering

\begin{tabular}{c|C{5.2em}C{5.2em}C{5.2em}}
\Xhline{2\arrayrulewidth}
 & \caregnn & \pcgnn & \gaga \\
\Xhline{\arrayrulewidth}
\clean & 48.02 & 55.62 & 21.68 \\
\Xhline{\arrayrulewidth}
random attributes & 78.18 & 77.81 & 25.70 \\
random edges & 79.64 & 81.40 & 39.33 \\
w/o pos. encoding & 86.12 & 86.98 & 40.21 \\
w/o degree & 87.33 & 88.06 & 38.40 \\
shared parameters & 87.76 & 86.66 & 39.57 \\
w/o candidates & 88.20 & 88.96 & 40.67 \\
random candidates & 88.25 & 88.38 & 39.13 \\
fixed budget & 87.28 & 88.66 & 42.17 \\
\Xhline{\arrayrulewidth}
\ours & \textbf{88.78} & \textbf{89.90} & \textbf{43.70} \\
\Xhline{2\arrayrulewidth}
\end{tabular}

\caption{Ablation studies of \ours on \gcop with \gcn as the surrogate model. We report misclassification rates (\%). \clean represents the misclassification rates of the victim models measured on the original clean graph. Overall, the original \ours shows the best attack performance, demonstrating the importance of each component of \ours.}
\label{tb:app_ablation}

\end{table}

\subsection{Efficiency Analysis}
\label{subsec:app_efficiency}
We measure the total training, inference, and running times, as well as the maximum GPU memory usage for each attack method on \gcop, \yelp, and \insr using \gcn as the surrogate model and \gaga as the victim model, as shown in Table~\ref{tb:app_efficiency}. Experiments on \gcop and \yelp were conducted using GeForce RTX 3090 24GB, while experiments on \insr were performed using NVIDIA T4 GPU. Note that as \tdgia and \cluster are algorithm-based methods, which require optimization for each attack, they only have inference times. We measure the maximum GPU memory usage using \texttt{torch.cuda.max\_memory\_allocated()}. We mainly compare \ours with \gnia, which is the best-performing baseline. Across all datasets, training \ours requires significantly less time than \gnia (e.g., 1,581 vs. 14,757 seconds on \gcop). The inference time of \ours is comparable to or slightly better than \gnia. Since \ours and \gnia are model-based methods which are generalizable to unseen nodes based on trained parameters, their inference times are significantly less than those of \tdgia and \cluster. The total runtime of \ours ranges from 11\% to 93\% of that required by baselines across different datasets. In terms of memory efficiency, \ours shows moderate GPU memory usage compared to baselines, while \gsquare encounters Out of Memory errors in all settings. \ours effectively narrows the search space through candidate selection and adopts efficient matrix operations to inject all attack nodes at once. Especially, the candidate selection mechanism significantly reduces the number of nodes involved in transformer computations of \ours by focusing solely on target nodes and a subset of their $K$-hop neighbors, rather than processing the entire graph. Although \ours requires more memory storage than algorithm-based methods such as \tdgia and \cluster, \ours operates much faster and shows superior attack performance compared to \tdgia and \cluster.

\begin{table}[t]
\small
\renewcommand{\arraystretch}{1.2}
\setlength{\tabcolsep}{0.25em}
\centering

\begin{tabular}{cc|C{7.2em}C{7.2em}C{7.2em}C{7.2em}|C{7.2em}}
    \Xhline{2\arrayrulewidth}
    & & \gnia & \tdgia & \cluster & \gsquare & \ours \\
    \Xhline{\arrayrulewidth}
    \multicolumn{1}{c}{\multirow{4}{*}{\gcop}} & Training Time & 14,757 $\pm$ 1,166 s & - & - & - & 1,581 $\pm$ 336.3 s \\
    & Inference Time & 123.4 $\pm$ 0.61 s & 9,874 $\pm$ 212.6 s & 3,686 $\pm$ 59.52 s & - & 119.2 $\pm$ 0.34 s \\
    & Total Runtime & 14,880 $\pm$ 1,166 s & 9,874 $\pm$ 212.6 s & 3,686 $\pm$ 59.52 s & - & 1,700 $\pm$ 336.0 s \\
    & Max. GPU Mem. Usage & 13.1 $\pm$ 0.00 GiB & 2.43 $\pm$ 0.00 GiB & 2.17 $\pm$ 0.00 GiB & OOM & 6.21 $\pm$ 0.01 GiB \\
    \Xhline{\arrayrulewidth}
    \multicolumn{1}{c}{\multirow{4}{*}{\yelp}} & Training Time & 7,574 $\pm$ 3,454 s & - & - & - & 1,365 $\pm$ 342.9 s \\
    & Inference Time & 56.85 $\pm$ 0.95 s & 1,397 $\pm$ 171.2 s & 1,158 $\pm$ 29.51 s & - & 52.84 $\pm$ 0.34 s \\
    & Total Runtime & 7,631 $\pm$ 3,455 s & 1,397 $\pm$ 171.2 s & 1,158 $\pm$ 29.51 s & - & 1,417 $\pm$ 342.9 s \\
    & Max. GPU Mem. Usage & 1.69 $\pm$ 0.14 GiB & 2.19 $\pm$ 0.00 GiB & 2.17 $\pm$ 0.00 GiB & OOM & 1.86 $\pm$ 0.00 GiB \\
    \Xhline{\arrayrulewidth}
    \multicolumn{1}{c}{\multirow{4}{*}{\insr}} & Training Time & 1,348 $\pm$ 304.3 s & - & - & - & 139.7 $\pm$ 14.4 s \\
    & Inference Time & 10.91 $\pm$ 0.08 s & 3,171 $\pm$ 86.16 s & N/A & - & 13.72 $\pm$ 0.46 s \\
    & Total Runtime & 1,359 $\pm$ 304.4 s & 3,171 $\pm$ 86.16 s & N/A & - & 153.4 $\pm$ 14.64 s \\
    & Max. GPU Mem. Usage & 3.54 $\pm$ 0.00 GiB & 3.85 $\pm$ 0.00 GiB & - & OOM & 3.51 $\pm$ 0.00 GiB \\
    \Xhline{2\arrayrulewidth}
\end{tabular}

\caption{Efficiency analysis results on \gcop, \yelp, and \insr using \gcn as the surrogate model and \gaga as the victim model. OOM indicates an Out of Memory error. N/A denotes that experiments were not completed within 5 days.}
\label{tb:app_efficiency}

\end{table}

\subsection{Complexity Analysis of \ours}
\label{subsec:app_complexity}
In this section, we analyze the computational complexity of \ours's forward pass. We do not consider the operations using the pretrained GNN encoder as these operations can be precomputed. The candidate selection process has a complexity of $O(|\mathcal{N}^{(K)}| + m)$ where $m := |\mathcal{T}|$ is the number of target nodes and $\mathcal{N}^{(K)}$ is a set of $K$-hop neighbors of the target nodes excluding the target nodes themselves. Note that $K$ is a tunable hyperparameter that allows us to control the trade-off between computational cost and the exploration range of candidate nodes. The complexity of the adversarial structure encoding step is $O((m + \alpha + \Delta)^2)$ for transformer-based encoding. The one-time graph injection has a complexity of $O(\Delta \cdot (m + \alpha + \Delta))$, which is absorbed into the adversarial structure encoding complexity. Therefore, \ours's overall complexity is:
\begin{equation}
O(|\mathcal{N}^{(K)}| + (m + \alpha + \Delta)^2)
\end{equation}
where $(m + \alpha + \Delta) \ll |\mathcal{V}|$. The computational cost of candidate selection does not grow fast since target nodes representing fraud gangs are often locally close, resulting in considerable overlap among neighboring nodes. For comparison, we analyze \gnia, the best-performing baseline. \gnia consists of attribute generation with $O(m + \Delta)$ complexity and edge generation with $O(\Delta \cdot (|\mathcal{N}^{(1)}| + m))$ complexity. Thus, \gnia's overall complexity is:
\begin{equation}
O(\Delta \cdot (|\mathcal{N}^{(1)}| + m))
\end{equation}
While \ours's complexity mainly depends on $|\mathcal{N}^{(K)}|$, which can be controlled through the hyperparameter $K$, the complexity of \gnia is determined by the product of $|\mathcal{N}^{(1)}|$ and node budget $\Delta$. We can see that the complexities of \ours and \gnia are affected by the dataset characteristics and chosen values of $K$ and $\Delta$. However, as shown in Appendix~\ref{subsec:app_efficiency}, \ours is significantly more efficient than \gnia in runtimes and memory usage in our experiments.

\subsection{More Examples of Visualization of Large-Scale Attack Cases}
\label{subsec:app_case}
\begin{figure}[t]
\centering
\includegraphics[width=\columnwidth]{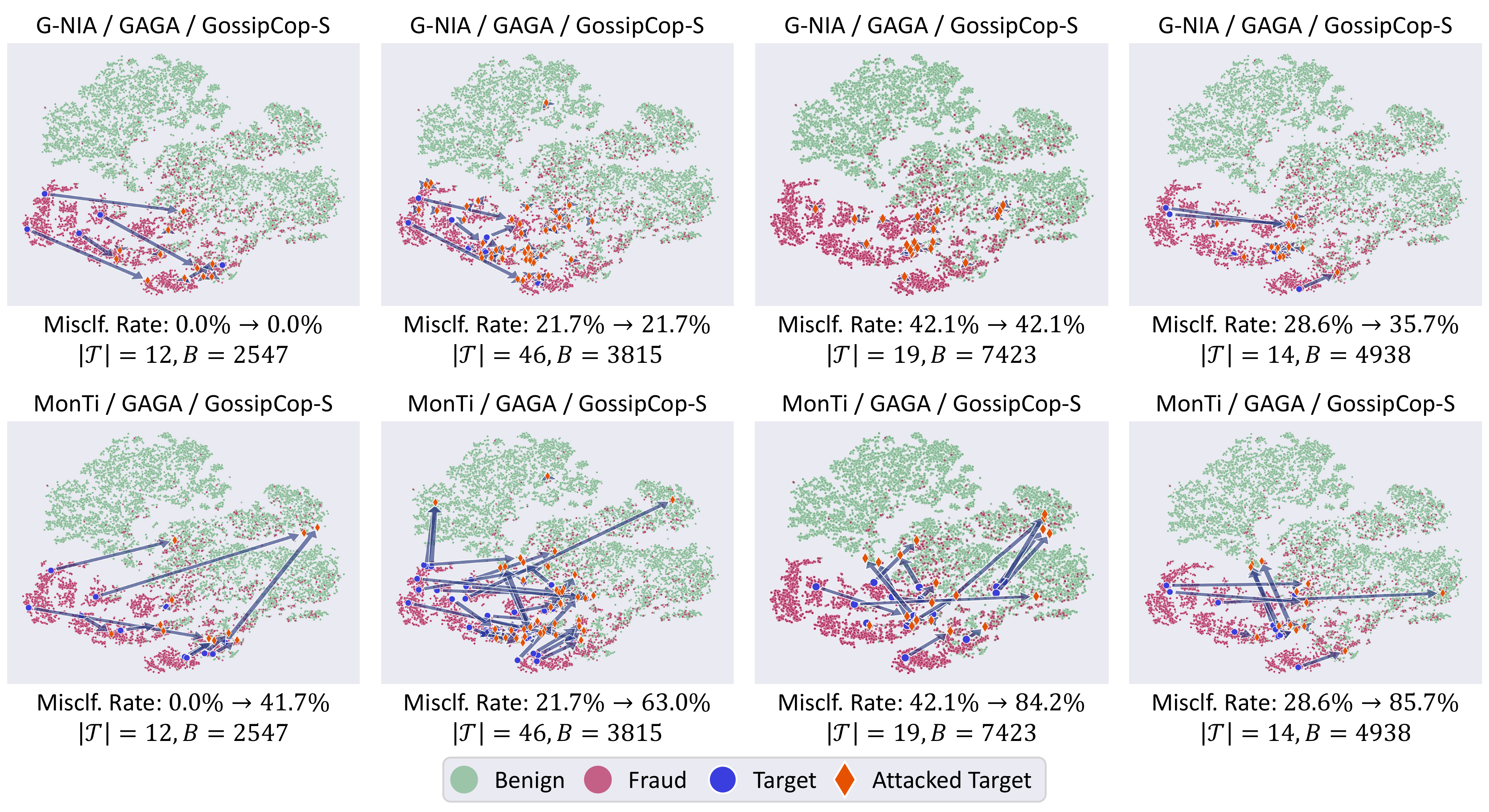}
\caption{The t-SNE visualization of the changes in the latent representations of target nodes computed by \gaga on \gcop, incurred by \gnia (Upper) and \ours (Lower). A blue circle and an orange diamond corresponding to the same target node are connected. For each figure, we also provide misclassification rates of \gaga for the corresponding target set before and after the attack, the size of the target set, and $B = \lvert \sN^{(1)} \cup \sT \rvert$. While \gnia induces only minor changes in the representations of target nodes, \ours significantly shifts the representations.}
\label{fig:app_case}
\end{figure}

Figure~\ref{fig:app_case} illustrates additional cases using the same settings as those of Figure 3 in the main paper: on \gcop using target sets with $B > 1000$, \gcn as the surrogate model, and \gaga as the victim model. We visualize the latent node representations computed by \gaga before and after the attack using t-SNE~\cite{tsne}. Blue circles indicate the representations of target nodes before the attack, and orange diamonds indicate those after the attack. A blue circle and an orange diamond that correspond to the same target node are connected. Below each t-SNE visualization, we provide misclassification rates of \gaga for the corresponding target set before and after the attack, the size of the target set, and $B = \lvert \sN^{(1)} \cup \sT \rvert$. These supplementary cases provide further validation of our findings. Specifically, \ours substantially shifts the representations of target nodes from the fraud region to the benign region, leading to a significant increase in the number of targets being misclassified. In contrast, \gnia induces only minor changes in target node representations, minimally increasing misclassification rates consequently. The first three figures in the upper row in Figure~\ref{fig:app_case} show that although \gnia shifts the representations of some target nodes, this does not lead to an increase in misclassification rates.

\end{document}